\documentclass[nonblindrev,copyedit]{informs3}

\usepackage{mathrsfs}
\usepackage{mathtools}
\usepackage{multirow}
\usepackage{graphicx,color}
\usepackage{xcolor}
\definecolor{DarkBlue}{rgb}{0,0.08,0.45}
\usepackage[backref = false, bookmarks, breaklinks=true, colorlinks = true, plainpages = false, citecolor = DarkBlue , urlcolor = DarkBlue, filecolor = DarkBlue, linkcolor = DarkBlue]{hyperref}
\usepackage{booktabs}
\usepackage{rotating}

\usepackage{bm}
\usepackage{enumitem}
\usepackage[top=0.8in, bottom=0.8in, left=1in,
right=1in]{geometry}
\usepackage{pgfplots}
\pgfplotsset{compat=1.14}

\DeclareMathOperator{\Var}{\rm{Var}}

\DeclareMathOperator*{\plim}{plim}

\theoremstyle{TH}
\newtheorem{assumption}{Assumption}
\newtheorem{proposition}{Proposition}
\newtheorem{theorem}{Theorem}
\newtheorem{lemma}{Lemma}
\newtheorem{corollary}{Corollary}

\theoremstyle{EX}

\newtheorem{example}{Example}

\usepackage[sort&compress]{natbib}
\bibpunct[, ]{(}{)}{,}{a}{}{,}
\def\bibfont{\small}

\newcommand*{\dif}{\mathop{}\!\mathrm{d}}
\newcommand{\PR}{\mathbb P}
\newcommand{\R}{\mathbb R}
\newcommand{\EX}{\mathbb E}

\newcommand{\xx}{x^\ast} 
\newcommand{\xa}{X_a} 
\newcommand{\sxa}{x_a} 
\newcommand{\xh}{X_h} 
\newcommand{\sxh}{x_h} 
\newcommand{\xin}{\hat{X}} 
\newcommand{\sxin}{\hat{x}} 
\newcommand{\xhl}{X_h^l} 
\newcommand{\sxhl}{x_h^l} 
\newcommand{\xhu}{X_h^u} 
\newcommand{\sxhu}{x_h^u} 
\newcommand{\pa}{p_a} 
\newcommand{\pas}{p_a^*}
\newcommand{\ph}{p'} 
\newcommand{\prin}{\hat{p}} 
\newcommand{\rup}{\overline{u}} 
\newcommand{\rlw}{\underline{u}}

\begin{document}
	\TITLE{\Large Algorithmic Decision-Making Safeguarded by Human Knowledge}
	
	\ARTICLEAUTHORS{
	\AUTHOR
	{Ningyuan Chen, Ming Hu, Wenhao Li}
	\AFF{
	Rotman School of Management, University of Toronto, Toronto, Ontario, Canada M5S 3E6 \\ \href{mailto:ningyuan.chen@utoronto.ca}{ningyuan.chen@utoronto.ca}, \href{mailto:ming.hu@rotman.utoronto.ca}{ming.hu@rotman.utoronto.ca}, \href{mailto:wwenhao.li@mail.utoronto.ca}{wwenhao.li@mail.utoronto.ca}}

%
	}
	\ABSTRACT{Commercial AI solutions provide analysts and managers with data-driven business intelligence for a wide range of decisions, such as demand forecasting and pricing. However, human analysts may have their own insights and experiences about the decision-making that is at odds with the algorithmic recommendation. In view of such a conflict, we provide a general analytical framework to study the augmentation of algorithmic decisions with human knowledge: the analyst uses the knowledge to set a guardrail by which the algorithmic decision is clipped if the algorithmic output is out of bound and seems unreasonable. We study the conditions under which the augmentation is beneficial relative to the raw algorithmic decision. We show that when the algorithmic decision is asymptotically optimal with large data, the non-data-driven human guardrail usually provides no benefit. However, we point out three common pitfalls of the algorithmic decision: (1) lack of domain knowledge, such as the market competition, (2) model misspecification, and (3) data contamination. In these cases, even with sufficient data, the augmentation from human knowledge can still improve the performance of the algorithmic decision.}
	\maketitle

\section{Introduction}\label{sec:intro}

The Russia-Ukraine war has sent a seismic wave to the energy market
and brought soaring gas prices under the spotlight.
Ideally, the optimal retail fuel price at the pump needs to take into account many factors, including 
the crude oil price, the transportation cost, the brand value, and the local competition.
Because of the complexity of the pricing problem, in practice, it is not surprising that the station managers would rely on some heuristics or simple rules (such as a constant markup over the cost) to set prices instead of using a sophisticated pricing algorithm.

This is no longer the case for many gas stations, especially those owned by a large corporation.
PDI Fuel Pricing\footnote{\url{https://www.pdisoftware.com/fuel-pricing-solutions/}}
sells software to gas station managers that helps them set fuel prices more intelligently using data analytics and machine learning.
It uses a wide range of data, including historical prices and demand, as well as competitors' prices  
and
claims to ``fine-tune your pricing strategy with live competitive insights allowing [managers] to react quickly to market conditions.''
Needless to say, machine learning algorithms can significantly improve profitability over the heuristic approach.

However, it is not hard to imagine scenarios when a human analyst or the station manager may not be fully convinced by the price prescribed by the algorithm, especially when the algorithm is a black box (typical for many machine learning algorithms)
and the prescribed price deviates from human intuition significantly. 
For example, when the algorithm recommends a price that is much higher than what would have been charged by the station manager, should the algorithmic decision be trusted over human knowledge?
On the one hand, the algorithm takes much more quantitative information as input than the human analyst, and the higher price could reflect the rising demand, a pattern in the data missed by the human analyst.
On the other hand, human knowledge may have relied on simple rules such as matching the price of another station around the corner.
Such price-matching heuristics may have worked well in the past.
When the decisions from the algorithm and human knowledge are in conflict, it can be hard for the analyst to make a call.

The problem faced by the station manager in the motivating example is prevalent.
Most business owners have realized the importance of the AI revolution and are willing to invest in it to improve business decision-making.
However, as AI algorithms become increasingly sophisticated, many firms have no choice but to outsource the standardized components in the decision-making process to commercial AI solutions.
These decisions, such as pricing and inventory management, have historically been made through human instincts and experiences. 
When human knowledge and the decision output by AI algorithms deviate significantly, firms face a similar dilemma to the station manager.

In this paper, we provide a general analytical framework to study practical problems in which humans and AI interact in the decision-making process.
Motivated by the gas station example, we consider an AI system prescribing a decision based on past data and some machine learning algorithms.
Based on the prescribed decision, the human analyst may set a guardrail using simple rules from the accumulated knowledge, experiences, or expertise.
More precisely, human knowledge is translated to a cap or floor, or both of the decision. 
That is, if the algorithmic decision violates the bounds, the human analyst may override it by clipping it to the imposed cap or floor. 
For example, the algorithm may recommend a retail price of \$5.10 per gallon. At the same time, human knowledge indicates, ``the price can't be higher than \$5.00 per gallon because the station around the corner is only charging \$4.80.''
As a result, the human analyst may set the final price to \$5.00.
Otherwise, if the recommended price is lower than \$5.00, then the algorithmic decision is followed.
In this interaction, AI is the main force behind the decision-making, while human knowledge serves as an auxiliary, safeguarding the algorithmic decision from prescribing unreasonably high prices. 
It is a fair representation of a considerable fraction of human-AI interaction in practice.

With the framework, we aim to answer the following research question:
When does human knowledge add value to AI decision-making?
Our first result is \emph{negative}:
human knowledge does not provide any benefit if 
(1) the algorithmic decision improves with more data, for example, when the mean squared error with respect to the optimal decision is diminishing, and
(2) human knowledge is not improving with more data. 
This result is somewhat expected: 
The guardrail prescribed by human knowledge can itself be treated as the pattern extrapolation of past data, albeit a simple and heuristic one.
If the algorithm can efficiently recognize and extrapolate the pattern better than the human, as many machine learning algorithms do, 
then it is unnecessary to augment the algorithmic decision with human knowledge under sufficient data.

The above result may sound intuitive, but it is derived in an ideal situation.
While it may be reasonable to assume that human knowledge does not improve constantly with more data, as human brains are generally unable to recognize complex patterns hidden in a large dataset, there are many caveats in applying commercial off-the-shelf AI systems to real-world applications as those algorithms may fail to satisfy condition (1) above.
In these cases, human knowledge can be used to augment the algorithmic decision.
In this study, we identify three such use cases within our framework 
and argue that, in these cases, rhetorically, the gas station manager should not completely delegate the pricing decision to the algorithm of PDI Fuel Pricing.
The three cases summarize the common pitfalls when making business decisions and trusting the algorithm blindly.
\begin{itemize}
	\item When the firm is in a \emph{competitive market}, and the algorithm fails to fully take into account the competitors' decision (due to incomplete data or algorithmic design), simple decision rules based on human knowledge, such as price matching, can improve the algorithmic decision. 
	This is not an uncommon setting. For example, PDI Fuel Pricing may not have direct access to the pricing data of other competing local stations unless they subscribe to some service as well. 
	We show that when a competitor sets a price near the Nash equilibrium, using the algorithmic price and matching it to the competitor's price when the algorithmic price is higher can improve the algorithmic decision.
	\item The algorithm may be susceptible to \emph{model misspecification}. In the pricing context, the algorithm may mistakenly treat the demand function as a linear function and recommend the optimal price based on the misspecified linear demand model. 
	On the other hand, the human analyst may simply observe which price generates the highest profit empirically in the past data without fitting or optimizing a model.
	This heuristic turns out to be quite robust to model misspecification.
	We show that human knowledge when used to safeguard the algorithm, can help mitigate the model misspecification and improve the profitability of the algorithmic decision.
	\item When the data fed into the algorithm are \emph{contaminated}, possibly due to the reporting or measurement error, then the relative insensitivity of the human knowledge to specific data points turns out to be a robust mechanism.
	Not surprisingly, the combination of human knowledge and the algorithmic decision can prevent the latter from being misguided by the contaminated data.
	We provide an analytical condition that characterizes the contamination level for human knowledge to prevail.
\end{itemize}
In all three cases, instead of an abstract AI system, we materialize the algorithmic decision and study linear regression, which allows us to concretely analyze the trade-off of safeguarding the regression output using simple rules. 
Linear regression is widely used and is representative of a more complex machine learning algorithm.
Such treatment allows us to provide technical conditions under which augmentation by human knowledge can improve the algorithmic decision.

This study contributes to the growing literature on human-AI collaboration. 
In some applications, it has been shown that AI lacks crucial human strengths such as domain knowledge and common-sense reasoning \citep{holstein2021designing,lake2017building,miller2019explanation}, 
which motivates the collaboration between AI and human experts on subjects including chess  \citep{case2018become,das2020leveraging}, healthcare  \citep{patel2019human,irvin2019chexpert,dai2021artificial}, criminal justice \citep{kleinberg2018human,grgic2019human}, education \citep{smith2012predictive,cheng2019explaining}, and public services \citep{chouldechova2018case,binns2018s}.
This study is motivated by business problems, and the human-AI interaction is uniquely defined by the context.
Below we review the literature closely related to this study.



\section{Related Literature}\label{sec:literature}
This research is broadly related to two streams of literature:
those papers providing conceptual or theoretical frameworks for human-AI collaboration and empirical papers documenting real-world interactions between AI and human analysts.
In the first stream, recent literature in computer science aims at the optimal integration of human and AI decisions \citep{madras2018predict,wilder2020learning,gao2021human,mozannar2020consistent,keswani2021towards,bansal2021most,bansal2019updates,donahue2022human,rastogi2022unifying,raghu2019algorithmic}. On the one hand, \citet{madras2018predict} propose a learning-to-defer framework in which the AI can choose to make decision by its own or just pass the task to the downstream human expert. 
The expert has information unavailable to AI and may make better decisions. 
Follow-up papers extend the framework to more complex settings, such as multiple experts \citep{keswani2021towards}, bandit feedback \citep{gao2021human}, joint optimization of the prediction algorithm and pass function \citep{wilder2020learning,mozannar2020consistent}.
On the other hand,  \citet{donahue2022human,rastogi2022unifying} consider a \emph{weighted average} aggregation of human and AI decisions and show conditions for human-AI complementarity in which the aggregated decision outperforms both individual decisions.
Recently, \citet{grand2022best} show that in the setting of sequential decision-making, the AI algorithm should be trained differently when a human analyst is involved.
Motivated by business applications,
our paper differs from this stream of works as we 
study a particular (not necessarily optimal) way to integrate the algorithmic and human decisions tailored to the application.
In the motivating example, the manager does not have access to the internal structure of the algorithm and cannot design a meta-algorithm to optimally instill her own knowledge into the algorithm.

Some recent studies in Operations Management analyze the human-AI interaction in a theoretical framework \citep{boyaci2020human,agrawal2018prediction,agrawal2019exploring,devericourt2022your,ibrahim2021eliciting,dai2021artificial}.
They focus on modeling the impact of AI-based predictions on the human decision-making process.
\citet{boyaci2020human} study the impact of AI predictions on human decision errors and the cognitive effort humans put into their decisions. The human has the cognitive flexibility to attend information from diverse sources but under limited cognitive capacity, while the AI only processes incomplete information but with great accuracy and efficiency. Through a rational inattention model, the authors show that AI prediction improves the overall accuracy of human decisions and reduces cognitive effort.
\citet{agrawal2018prediction} consider the human analyst aiming to maximize the utility which depends on their decision and the uncertain state. The state can be predicted accurately by the AI algorithm. But the human needs to learn the utility function. The authors show that AI prediction generally complements the human effort but could be a substitute in some cases. 
\citet{devericourt2022your} consider the human-AI interactions in a sequential setting in which the analyst gradually learns the accuracy of the AI algorithm through a sequence of tasks. Since the analyst can override AI and never actively explores the AI accuracy, the analyst may never know whether AI outperforms herself at the end of the day. The authors provide explanations for the coexistence of AI and humans, even if one actually outperforms the other.
\citet{dai2021artificial} use a theoretical framework to analyze a physician's decision with regard to whether to use AI when prescribing a treatment. They find that physicians may intentionally avoid using AI, even when AI can help mitigate clinical uncertainty because doing so increases their liability when adverse patient outcomes occur.
Our paper differs from these papers in modeling the human decision-making process. 
In our model, we assume the human analyst aims to directly safeguard the AI decisions using intuition and expertise.
We focus on whether the integration improves the raw AI output.

Empirical evidence shows that human knowledge can still improve AI systems, even though the latter have access to big data and computational resources
\citep{van2010ordering,campbell2011market,phillips2015effectiveness,karlinsky2019automating,kesavan2020field,Liu2022Alibaba,sun2022predicting}.
For example, in the context of inventory replenishment, \citet{van2010ordering} find that store managers often modify the algorithmic recommendation from an automated replenishment system. 
\citet{kesavan2020field} use the data from a field experiment to investigate the merchant's modification of the advice from a data-driven central-planning system.
They find that the merchant's modification reduces the overall profitability but improves the profit for growth-stage products whose historical data are limited.
\citet{Liu2022Alibaba} conduct a field experiment to compare the inventory replenishment strategies of human buyers and AI algorithms. They find the algorithm outperforms human buyers in terms of reducing out-of-stocks rates and inventory rates.
The most related empirical works to our study are \citet{ibrahim2021eliciting, fogliato2022case}. \citet{ibrahim2021eliciting} show how to exploit the human domain knowledge to improve the AI predictions for surgery duration. Particularly, they suggest inputting the human adjustment (so-called private information adjustment in the paper), instead of the human direct forecast, into the prediction algorithm. Their work conveys a message similar to ours: even the human predictions are less accurate than AI, they can still help boost AI performance.
\citet{fogliato2022case} investigate human-AI collaboration in the context of child maltreatment hotline screening. Due to the technical glitch caused by incorrect input, the AI may incorrectly predict the risk score in some cases. They find that human analysts are more likely to override AI recommendations when AI makes a mistake. The work shows that humans can augment the algorithmic decision when the algorithm exhibits defects in real-world applications. 
Our work provides a theoretical framework to complement the empirical evidence provided in the above papers and analyzes the situations when the human augmentation of algorithmic outputs is beneficial.

Another stream of the related empirical literature is ``judgmental adjustment of statistical forecasts'' (see \citealt{arvan2019integrating,lawrence2006judgmental} for a review). These studies consider the demand forecast problem in supply chain management. The human analyst is allowed to adjust the forecasts generated by an algorithm. The adjustment can improve the accuracy when the algorithmic forecast is deficient or the human has important domain knowledge that is unavailable to AI \citep{lawrence2006judgmental}. Several empirical studies aim to investigate the effect of the direction and magnitude of the adjustment on accuracy \citep{fildes2009effective,davydenko2013measuring,baker2021maximizing}. 
However, the benefit of such adjustments may be highly context-dependent \citep{khosrowabadi2022evaluating}. Although we consider a general decision-making problem, our work can also contribute to this literature by providing an analytical framework to characterize when the adjustment adds value to the algorithmic forecast.

Finally, we notice some recent works focusing on how to design user-friendly AI algorithms which the human analyst can easily understand and follow. \citet{bastani2018interpreting} construct extracted decision trees to interpret complex, black-box AI models and summarize their reasoning process. Applied to the diabetes risk prediction problem, the proposed algorithm produces more accurate interpretations than baseline algorithms. \citet{bastani2021learning} propose a reinforcement-learning algorithm for inferring interpretable tips to help workers improve their performance in sequential decision-making tasks. Through a virtual kitchen-management game, they show that the algorithm improves workers' performance. \citet{dietvorst2018overcoming} find that giving the human analyst some control over the AI output can reduce human's aversion to algorithms. 

\section{An Analytical Framework for Human-Safeguarded Algorithmic Decisions}\label{sec:general-loss}
In the retail fuel example, the gas station manager intends to set prices to maximize the profit.
The objective can be viewed more generally as the minimization of the loss in comparison to the optimal price.
In this section, we consider a general problem that an analyst intends to minimize a loss function $l(\cdot): \R\to\R$, which measures the loss due to the deviation from the optimal decision $\xx$, e.g., the profit loss due to making suboptimal operations or pricing decisions.
The loss function may represent the operational cost or the expected negative profit.
The analyst may not know the form of the loss function exactly and seeks the help from AI algorithms.
We do not impose any structure of the loss function but make the following mild assumptions.
\begin{assumption}\label{ass:general-loss}
	Suppose the loss function $l(\cdot)$ satisfies: (\romannumeral1) $l(x) $ is nonnegative;
	(\romannumeral2) $l(x)$ is quasiconvex with minimizer $\xx$.
\end{assumption}
Part (\romannumeral1) of Assumption \ref{ass:general-loss} is without loss of generality as the loss function can be shifted up by a constant of $|l(\xx)|$.
We first give two examples that will serve as running examples throughout the rest of the paper.
The two examples are intended to give the context of the loss function and demonstrate the generality of Assumption~\ref*{ass:general-loss}.
\begin{example}[Predictive Analytics: Prediction]\label{exp:prediction}
	If a firm intends to forecast a quantity, for example, the demand in the next season, 
	then the firm's problem can be cast as a prediction problem:
	The goal is to minimize the loss function $l(x)=(x-\xx)^2$, where $\xx$ is the actual value of the quantity of interest.
\end{example}
\begin{example}[Prescriptive Analytics: Pricing]\label{exp:pricing}
	Sophisticated algorithms such as online learning has been widely used in pricing (see, e.g., \citealt{den2022dynamic,keskin2022data}).
	When a new product is launched to the market, the retailer needs to set its price $x$. The goal of the retailer is to maximize the profit, which is the product of the profit margin $x-c$, where $c$ is the marginal cost and demand, i.e., $\pi(x)=(x-c) f(x)$. Denote $\xx$ by the optimal price. The retailer knows the marginal cost, but does not know the demand function $f(x)$ nor the optimal price.
	The loss function can be written as $l(x)=\pi(\xx)-\pi(x)$. 
	If the profit function $\pi(\cdot)$ is unimodal, then the loss function satisfies Assumption~\ref{ass:general-loss}. 
\end{example}
We next introduce the algorithmic decision and human knowledge into the framework.


\textbf{Algorithmic decision.} To accommodate a wide range of algorithms, we simply use a generic random variable $\xa$ to represent the decision.
The randomness may come from the randomness in the historical data or the randomization of the algorithm itself.
The performance of the algorithmic decision is thus evaluated by $\EX[l(\xa)]$.

\textbf{Human knowledge.} 
We focus on human knowledge in the form of a guardrail.
That is, the human analyst forms a belief with an upper bound on the optimal decision, based on her domain knowledge and experiences.
We use a random variable $\xh$ to denote the upper bound.
In Example~\ref*{exp:pricing}, $\xh$ could be interpreted as a price cap manually imposed by the retailer.
Note that unlike the algorithmic decision, $\xh$ is usually not data-dependent and tends to be stable, although we allow it to be random and correlated with $\xa$.
We use the upper bound as a form of domain knowledge due to two reasons.
First, it is common for human brains to perceive uncertainty in terms of intervals and worst-case scenarios. 
The notion is closely related to confidence intervals in statistics that have shaped how human's belief is formed.
Second, compared to point estimators, the notion we propose is more flexible and allows for different confidence levels. 

To keep the framework general, we do not specify how $\xa$ and $\xh$ are generated.
For $\xa$, it may be output by a machine learning algorithm deployed by the analyst
or a black-box commercial software as mentioned in the fuel-pricing example in the introduction.
The complexity of the algorithm may vary, e.g., linear regression versus neural networks.
The random variable $\xa$ can fully capture the wide range of scenarios.
For $\xh$, although itself may not represent a sensible decision, it may serve as a safeguard distilled from the accumulated knowledge of the human analyst.
Depending on the conservativeness and the risk preference of the analyst, $\xh$ may have different values.
For instance, in Example~\ref*{exp:prediction}, $\xh$ may roughly be the upper confidence bound of the targeted quantity with various confidence levels.


\textbf{Human-safeguarded algorithmic decision.} 
We consider a simple yet pervasive approach to integrate the algorithmic decision and human knowledge.
The human analyst safeguards the algorithmic decision by using 
\begin{equation}\label{eq:integration}
\xin \triangleq \min\{\xa, \xh\}.
\end{equation}
This is a rather natural step: 
the analyst follows the algorithmic decision if the upper bound is not violated;
otherwise, the upper bound is used.
Consider the example mentioned in the introduction (a special case of Example~\ref*{exp:pricing}), $\xa$ is the price output by PDI Fuel Pricing; $\xh$ is the price cap imposed by the station manager. 
The safeguarded algorithmic decision takes the minimum of the two, guaranteeing that the price output by the algorithm does not exceed the price cap.
This type of augmentation also captures the interaction between autonomous drones and vehicles and their human overseers \citep{berger2022}, in which the human overseer needs to step in and override the algorithm when the system encounters an unexpected situation.

Note that neither the algorithm nor the human analyst has access to the optimal decision $\xx$.
If $\xh\ge \xx$ almost surely, i.e., the upper bound provided by the human belief is indeed always larger than the \emph{true} optimal decision, then 
we can show that the safeguarded decision $\xin$ outperforms the raw algorithmic decision $\xa$, i.e., $\EX[l(\xin)]\le \EX[l(\xa)]$.
To see this, note that 
\begin{equation*}
\EX[l(\xin)]-\EX[l(\xa)]=\int_{\xx}^{\infty} \int_{x_h}^{\infty} (l(x_h)-l(x_a)) f(x_a,x_h) dx_a dx_h \leq 0,
\end{equation*}
where $f(\cdot,\cdot)$ represents the joint PDF and the inequality follows from $l(\xa) \ge l(\xh)$ for $\xa \ge \xh \ge \xx$.

The condition $\xh\ge \xx$, however, cannot be guaranteed, because the human analyst does not have precise information about $\xx$.
On one hand, when an unnecessary guardrail $\xh < \xx$ is imposed, the performance of $\xa$ is hurt for $\xa \in [\xh,\xx]$. 
In other words, if the suggested $\xh$ by the analyst is too aggressive, then $\xh< \xx$ is likely to happen and the human belief ends up clipping the algorithmic output $\xa$ for too many possible scenarios, even though the latter may accurately achieve the true optimal decision $\xx$.
Such an unnecessary guardrail inevitably introduces a significant downward bias and may cause the safeguarded algorithmic decision to be worse.
The faulty human knowledge leads to an additional cost to the AI decision.
On the other hand, one may argue that $\xh$ can be a sufficiently large number, so that $\xh\ge \xx$
always holds.
However, in this case, the human knowledge is almost useless in the process, as it does not provide a meaningful upper bound.
The improvement by the human augmentation, if any, is going to be minimal.
This is the result of an overly conservative human belief.
The observation highlights the impact of aggressive/conservative human augmentation. 
In the next proposition, we quantify the benefit of human augmentation.



\begin{proposition}[Conditions for beneficial human augmentation] \label{prop:loss-one-side-up}
	Suppose Assumption~\ref{ass:general-loss} holds.
	\begin{enumerate}[label={(\roman*)}]
		\item  We can quantify the benefit of human augmentation by 
		\begin{equation}\label{equ:loss-closed}
		\EX[l(\xa)]-\EX[l(\xin)]=\EX[(l(\xa)-l(\xh)) \mathbb{I} (\xh \leq \xa)].
		\end{equation}
		\item  A sufficient condition for beneficial augmentation $\EX[l(\xin)] \le \EX[l(\xa)]$ is 
		\begin{equation}\label{equ:loss-sufficient-low}
		\EX[l(\xa) \mathbb{I}(\xa > \xx,\xh \le \xx)] \ge \EX[l(\xh)\mathbb{I}(\xh \le \xx)].
		\end{equation}
		\item  A necessary condition for beneficial augmentation $\EX[l(\xin)] \le \EX[l(\xa)]$ is
		\begin{equation}\label{equ:loss-sufficient-high}
		\EX[l(\xa) \mathbb{I}(\xa \ge \xx)]  \ge \EX[l(\xh)\mathbb{I}(\xh \le \xx,\xa > \xx)].
		\end{equation}
	\end{enumerate}
\end{proposition}

Next, we interpret the result of Proposition~\ref{prop:loss-one-side-up}.
In \eqref{equ:loss-closed}, the benefit of human augmentation depends on the performance of the algorithmic decision $l(\xa)$ and human knowledge $l(\xh)$ on the event that the guardrail takes effect, i.e., $\mathbb{I} (\xh \leq \xa)$.
This is intuitive because the analyst counts on their knowledge to improve the algorithmic decision when it looks ``unreasonable.''
Conditions \eqref{equ:loss-sufficient-low} and \eqref{equ:loss-sufficient-high} are easier to interpret when $\xa$ and $\xh$ are independent, although we allow $\xa$ and $\xh$ to be dependent. For example, suppose the human knowledge $\xh$ is data-independent.
Then, \eqref{equ:loss-sufficient-low} and \eqref{equ:loss-sufficient-high} are reduced to, respectively,
\begin{align}
\EX[l(\xa) \mathbb{I}(\xa > \xx)] \ge \EX[l(\xh)\mathbb{I}(\xh \le \xx)]/\PR(\xh \le \xx) 
\label{equ:loss-sufficient-low-in}\\
\EX[l(\xa) \mathbb{I}(\xa \ge \xx)] \ge \EX[l(\xh)\mathbb{I}(\xh \le \xx)] \PR(\xa > \xx)  . \label{equ:loss-sufficient-high-in}
\end{align} 
The left-hand sides of \eqref{equ:loss-sufficient-low-in}
and \eqref{equ:loss-sufficient-high-in} measure the performance of the algorithmic decision. In particular, \eqref{equ:loss-sufficient-low-in} says that it is beneficial to safeguard the algorithmic decision when the right-hand side $\EX[l(\xh)|\xh\le \xx]$ is small enough.
In other words, the conditional expected loss does not explode when the human makes mistakes and imposes an overly aggressive bound ($\xh\le \xx$). Moreover, the necessary condition \eqref{equ:loss-sufficient-high-in}, which is weaker than \eqref{equ:loss-sufficient-low-in}, implies not to safeguard the algorithmic decision when the expected loss ($\EX[l(\xh)\mathbb{I}(\xh \le \xx)]$) incurred by human belief is over a certain amount.


It is easier to check whether the human augmentation is beneficial using \eqref{equ:loss-sufficient-low-in} and \eqref{equ:loss-sufficient-high-in}, than directly comparing \eqref{equ:loss-closed} with zero. This is because \eqref{equ:loss-closed} depends on the joint distribution of $(\xh,\xa)$ and the values of $l(\xh)$ and $l(\xa)$. However, in practice, the analyst may have collected the data in the past decision epochs during which one of the algorithmic and the human decisions has been applied and their realized losses have been observed. It may not be the case that the realized $l(\xh)$ and $l(\xa)$ can be observed simultaneously. While the conditions \eqref{equ:loss-sufficient-low-in} and \eqref{equ:loss-sufficient-high-in} only require the marginal distributions of $\xh$ and $\xa$ to be observed, which allows the analyst to evaluate whether human augmentation is effective in a data-driven manner.

Furthermore, we show the tightness of the sufficient condition \eqref{equ:loss-sufficient-low-in} relative to the necessary condition \eqref{equ:loss-sufficient-high-in}.
The right-hand sides of both \eqref{equ:loss-sufficient-low-in} and \eqref{equ:loss-sufficient-high-in}
has the common term $\EX[l(\xh)\mathbb{I}(\xh \le \xx)]$. 
The residual multipliers $1/\PR(\xh \le \xx)$ and $\PR(\xa \ge \xx)$ in \eqref{equ:loss-sufficient-low-in} and \eqref{equ:loss-sufficient-high-in}
tend to be constant even with increasing data sizes because 
in the former, human knowledge usually does not scale with big data,
while in the latter, for unbiased algorithmic decisions, $\PR(\xa \ge \xx)\approx 1/2$.
So the sufficient and necessary conditions tend to only differ by a constant factor. 
In the next example, we show that the sufficient condition in Proposition~\ref{prop:loss-one-side-up} cannot be improved even when the likelihood $\PR(\xh \le \xx)$ diminishes, and the right-hand side of \eqref{equ:loss-sufficient-low-in} cannot be relaxed to a constant multiplying $\EX[l(\xh)\mathbb{I}(\xh \le \xx)]$.
As a result, the condition tends to be tight.

\begin{example}[Tightness of the sufficient condition \eqref{equ:loss-sufficient-low-in}]\label{prop:loss-counter-example}
	Suppose $l(x)=(x-\xx)^2$, $\xa \sim N(\xx,\sigma^2)$, and $\xh$ satisfies
	\begin{equation}\label{equ:loss-counter-example}
	\PR(\xh=x)=\left\{
	\begin{array}{ll}
	1-\epsilon & \text{if} \ x=\infty,\\
	\frac{3 \epsilon}{(x-\xx)^4}  & \text{if} \ x \le \xx-1.\\
	\end{array}
	\right.
	\end{equation}
	Then for any $a \geq 1/4$, if $\epsilon \in (0, \sigma^2/(6a))$, $\xx<1$, and $\sigma^2<3/2$, we have $\EX[l(\xa) \mathbb{I}(\xa \ge \xx)] \ge a\EX[l(\xh)\mathbb{I}(\xh \le \xx)]$, but $\EX[l(\xin)] > \EX[l(\xa)]$.
\end{example}
Example~\ref{prop:loss-counter-example} shows the sufficient condition in Proposition~\ref{prop:loss-one-side-up} no longer holds if $\PR(\xh \le \xx)$ in \eqref{equ:loss-sufficient-low-in} is replaced by a constant.  
To better understand \eqref{equ:loss-sufficient-low-in} and \eqref{equ:loss-sufficient-high-in}, we show the conditions for Example~\ref{exp:prediction} (predictive analytics: prediction).
\begin{example}[Conditions for beneficial augmentation for the prediction problem]
	Consider historical samples $Z_1,\ldots, Z_n \sim N(\xx,\sigma^2)$. 
	The AI algorithm estimates $\xx$ by the sample mean $\xa=\frac{1}{n} \sum_{i=1}^n Z_i$, which follows the distribution $N(\xx,\sigma^2/n)$. 
	By Proposition~\ref{prop:loss-one-side-up}, if the upper bound derived from the human belief satisfies $\EX[(\xh-\xx)^2| \xh \le \xx] \le \sigma^2/(2n)$, then the augmentation improves the algorithmic decision. 
	On the other hand, if $\EX[(\xh-\xx)^2 \mathbb{I}(\xh \le \xx)] \ge \sigma^2/n$, then the augmentation is not beneficial. 
\end{example}



Proposition~\ref{prop:loss-one-side-up} provides us with an analytical framework to analyze the benefit of augmentation.
Based on the framework, we can show that there is an optimal level of safeguard when the human belief is deterministic.
\begin{corollary}[Optimal safeguard]\label{cor:loss-reduction}
	Suppose $\xh=x_h$ is a constant. Then the benefit of augmentation $\EX[l(\xa)]-\EX[l(\xin)]$ is unimodal in $x_h$, i.e., it increases when $x_h \le \xx$ and decreases when $x_h \ge \xx$.
\end{corollary}
Corollary \ref{cor:loss-reduction} holds under the condition that the human belief is deterministic. In this case, although the human analyst imposes an upper bound, it is the best to equate it to the true optimal decision $\xx$, i.e., a buffer is not necessary.
Of course, the corollary cannot provide a guidance for the human analyst to select the optimal bound, because $\xx$ is not accessible.
It does show the trade-off between conservative/aggressive guardrails.

Symmetrically, we can derive similar results when the guardrail derived from the human domain knowledge takes the form 
of a lower bound.
\begin{corollary}[Safeguarded by a lower bound] \label{prop:loss-one-side-low}
	Consider $\xin=\max\{\xa,\xh\}$. We have (\romannumeral1)
	$
	\EX[l(\xa)]-\EX[l(\xin)]=\EX[(l(\xa)-l(\xh)) \mathbb{I} (\xh \geq \xa)]$.
	(\romannumeral2) A sufficient condition for $\EX[l(\xin)] \le \EX[l(\xa)]$ is 
	\begin{equation}\label{equ:loss-sufficient-low-low}
	\EX[l(\xa) \mathbb{I}(\xa < \xx,\xh \ge \xx)] \ge \EX[l(\xh)\mathbb{I}(\xh \ge \xx)].
	\end{equation}
	And (\romannumeral3) a necessary condition for $\EX[l(\xin)] \le \EX[l(\xa)]$ is
	\begin{equation}
	\EX[l(\xa) \mathbb{I}(\xa \le \xx)]  \ge \EX[l(\xh)\mathbb{I}(\xh \ge \xx, \xa < \xx)].
	\end{equation}
\end{corollary}

Next we extend the results by two-sided bounds.
In particular, suppose the human analyst imposes both lower and upper bounds on the algorithmic decision.
For example, in the motivating example in the introduction, 
the station manager may propose a range for the retail price: the markup has to be between $\$0.10/L$ and $\$0.30/L$, regardless of the recommendation of the algorithm.
Mathematically, 
the human belief is translated to an interval $[\xhl,\xhu]$. The algorithmic decision $\xa$ is then projected onto the interval, i.e., $\xin=\min\{\max\{\xa,\xhl\},\xhu\}$.
Proposition~\ref{prop:loss-two-side} characterizes the benefit of such augmentation.

\begin{proposition}[Benefit of safeguarding using a two-sided bound] \label{prop:loss-two-side}
	Suppose Assumption \ref{ass:general-loss} holds. 
	\begin{enumerate}[label={(\roman*)}]
		\item The benefit of the human safeguard by a two-sided bound is
		\begin{equation*}
		\EX[l(\xa)]-\EX[l(\xin)]=\EX[(l(\xa)-l(\xhl)) \mathbb{I} (\xa \leq \xhl)]+\EX[(l(\xa)-l(\xhu)) \mathbb{I} (\xa \geq \xhu)].
		\end{equation*}
		\item  A sufficient condition for $\EX[l(\xin)] \le \EX[l(\xa)]$ is 
		\begin{align}\label{equ:loss-sufficient-low-two}
		\EX[l(\xa) \mathbb{I}(\xa \ge \xx,\xhu \le \xx)] &+\EX[l(\xa) \mathbb{I}(\xa \le \xx,\xhl \ge \xx)] \notag\\
		&\ge \EX[l(\xhu)\mathbb{I}(\xhu \le \xx)]+\EX[l(\xhl)\mathbb{I}(\xhl \ge \xx)]. 
		\end{align}
		\item A necessary condition for $\EX[l(\xin)] \le \EX[l(\xa)]$ is
		\begin{align}\label{equ:loss-sufficient-high-two}
		\EX[l(\xa)] \ge \EX[l(\xhu)\mathbb{I}(\xhu \le \xx \le \xa)] +\EX[l(\xhl)\mathbb{I}(\xa \le \xx \le \xhl)].
		\end{align}
	\end{enumerate}
\end{proposition}
If the bounds satisfy $\PR(\xhl \le \xx \le \xhu)=1$,
i.e., they always enclose the actual optimal decision,
then the safeguard always improves the algorithmic decision.
When this condition fails, 
\eqref{equ:loss-sufficient-low-two} and \eqref{equ:loss-sufficient-high-two} imply conditions to check whether to safeguard the algorithmic decision. 
Intuitively, the safeguard is beneficial when the loss incurred by the interval not covering $\xx$ is relatively small compared to the loss of the algorithmic decision. 

One can see that Proposition~\ref{prop:loss-two-side} reduces to Proposition~\ref{prop:loss-one-side-up} and Corollary~\ref{prop:loss-one-side-low} when $\xhl=-\infty$ or $\xhu=\infty$. 
We point out that the conditions for two-side bounds are weaker than the conditions for the one-sided bound. If $\xhu$ satisfies \eqref{equ:loss-sufficient-low} and $\xhl$ satisfies \eqref{equ:loss-sufficient-low-low}, then $[\xhl,\xhu]$ satisfies \eqref{equ:loss-sufficient-low-two}. But the reverse is not true. So the two-side conditions allow the human to make more mistakes in one side as long as the loss can be compensated by the other.

\subsection{Covariate Information}\label{sec:covariate}
So far, we have considered a simple model that the environment does not provide any covariate information at the specific decision epoch. However, 
in many data-driven decision-making problems, the analyst may observe additional covariate information $W$ and hence the optimal decision $\xx$ can depend on such covariate information. 
Upon observing the covariates, the algorithm outputs a decision $\xa(W)$.
In the prediction problem (Example~\ref{exp:prediction}), one can think of $W$ as the new input to the prediction algorithm such as weather conditions.
In the pricing problem (Example~\ref{exp:pricing}), $W$ may represent the available side information about the market to assist the choice of the optimal price. 
For example, PDI Fuel Pricing would take into account the crude oil price, which is a major cost component,  
to determine the retail gas price.  
In this case, the crude oil price is changing over time and can be considered as part of the covariate information.

After receiving the algorithmic recommendation, the human analyst comes up with a bound $[\xhl(W),\xhu(W)]$ to safeguard it.
That is, $\xin(W)=\min\{\max\{X(W),\xhl(W)\},\xhu(W)\}$. 
Note that in many cases the human domain knowledge may not be sophisticated enough to adapt to a specific covariate $W$.
In such cases, $\xhl$ and $\xhu$ do not depend on $W$, which is also covered by our framework.

When the covariate information is available, the loss function $l(x,w)$ depends on both the decision and the covariate.
We impose the following assumption, in parallel to Assumption~\ref{ass:general-loss}.
\begin{assumption}\label{ass:general-loss-covariate}
	Assume the loss function $l(x,w)$ satisfies the following conditions.
	\begin{enumerate}[label={(\roman*)}]
		\item For any decision $x \in \R$ and any covariate $w \in \R^d$, $l(x,w) \ge 0$.
		\item For any covariate $w \in \R^d$, $l(\cdot,w)$ is quasi-convex with minimizer $\xx(w)$.
	\end{enumerate}
\end{assumption}
Next, we characterize the benefit of human augmentation in the presence of covariate information by generalizing Proposition~\ref{prop:loss-one-side-up}.
Note that $\xh$, $\xa$, and $\xx$ all depend on (and are correlated with) $W$.
We omit the dependence for the readability.
\begin{proposition}[Benefit of human augmentation with covariate information] \label{prop:loss-two-side-cov}
	Suppose Assumption \ref{ass:general-loss-covariate} holds.
	\begin{enumerate}[label={(\roman*)}]
		\item  The benefit of human augmentation is
		\begin{align}\label{equ:loss-two-closed-cov}
		\EX[l(\xa,W)]-\EX[l(\xin,W)]
		&=\EX[(l(\xa,W)-l(\xhl,W)) \mathbb{I} (\xa \leq \xhl)] \notag\\
		&\quad+\EX[(l(\xa,W)-l(\xhu,W)) \mathbb{I} (\xa \geq \xhu)].
		\end{align}
		\item  A sufficient condition for $\EX[l(\xin,W)] \le \EX[l(\xa,W)]$ is 
		\begin{align}\label{equ:loss-sufficient-low-two-cov}
		\EX[l(\xa,W) \mathbb{I}(\xa \ge \xx,\xhu \le \xx)] &+\EX[l(\xa,W) \mathbb{I}(\xa \le \xx, \xhl \ge \xx)]  \notag\\
		&\ge \EX[l(\xhu,W)\mathbb{I}(\xhu \le \xx)]+\EX[l(\xhl,W)\mathbb{I}(\xhl \ge \xx)]. 
		\end{align}
		\item A necessary condition for $\EX[l(\xin,W)] \le \EX[l(\xa,W)]$ is
		\begin{align}\label{equ:loss-sufficient-high-two-cov}
		\EX[l(\xa,W) ]
		\ge \EX[l(\xhu,W)\mathbb{I}(\xhu \le \xx,\xa \ge \xx)]+\EX[l(\xhl,W)\mathbb{I}(\xhl \ge \xx,\xa \le \xx)] .
		\end{align}
	\end{enumerate}
\end{proposition}
When $W$ is a constant, Proposition~\ref{prop:loss-two-side-cov} reduces to the special case of Proposition~\ref{prop:loss-two-side}.
As expected, the conditions in Proposition~\ref{prop:loss-two-side-cov}  are more involved, although they follow a similar form to Proposition~\ref{prop:loss-two-side}.
To explain the intuition, we adopt the following example.

\begin{example}[Linear regression]\label{example:MSE-LR}
	Linear regression is a special case of the prediction problem (Example~\ref{exp:prediction}) with covariates.
	Suppose the loss function is $l(x,w) = (x-w^\top\beta)^2$ for some unknown coefficient $\beta$.
	As a result, the optimal decision is $\xx(W)=W^\top \beta$. 
	Using the least squares estimator $\hat\beta$, the algorithmic output is $\xa(W) = W^\top \hat\beta$.
	In the necessary condition \eqref{equ:loss-sufficient-high-two-cov},
	the left-hand side is the mean-squared error (MSE) of the least-squares estimation, which typically converges to zero at the rate of $1/n$, where $n$ is the sample size.
	In this case, for the human augmentation to outperform the algorithm, the right-hand side of \eqref{equ:loss-sufficient-high-two-cov} should diminish at the same or a faster rate.
	It is only possible if the bounds derived from human belief, $\xhl$ and $\xhu$, have diminishing MSEs
	$l(\xhl,W)$ and $l(\xhu,W)$, or they almost always sandwich the optimal $x^*$, i.e., $ \xhl \le x^*\le \xhu$.
	Both of the above two requirements set impractically high bars for the human domain knowledge.
\end{example}


From the example, we see that AI incurs diminishing loss as it gathers more data. 
In a data-rich environment, it appears that the human domain knowledge is not likely to improve AI. 
However, Example~\ref*{example:MSE-LR} does not reflect one of the major reasons why human domain knowledge may be helpful:
algorithms designed for general purposes sometimes ignore practical factors in the training dataset, such as contamination, model misspecification, and data errors. 
In the following sections, we provide examples to show that even if AI has a large amount of data, the human knowledge can still play an important role and contribute to the decision-making.

\section{Three Use Cases on Beneficial Human Augmentation}
In this section, we provide three concrete use cases when the safeguard derived from human knowledge can indeed improve algorithmic decisions even with large data, despite the potential limitation of human knowledge illustrated in Example~\ref{example:MSE-LR}.
We first provide a summary of the three use cases as follows:
\begin{itemize}
    \item In Section~\ref{sec:compe-pricing}, we consider a pricing problem (Example~\ref{exp:pricing}) under competition. We show that when the algorithm fails to take into account the competitive environment the pricing problem resides in, 
    simple human augmentation like price matching can improve the algorithmic decision.
    \item In Section~\ref{sec:mis}, we consider a pricing problem when the algorithm misspecifies the demand function. We show that using empirical observations (in particular, setting a price interval  using the historical price range that contains the highest profit in the past) can improve the performance of the algorithm. 
    \item In Section~\ref{sec:conta-LR}, we show that in prediction problems (Example~\ref{exp:prediction}), the human knowledge can serve as a robust mechanism to limit the damage due to data contamination and thus improve the algorithm's performance.
\end{itemize}

\subsection{Pricing Algorithm under Competition}\label{sec:compe-pricing}
In this section, consider the pricing problem of a focal firm when there is a competitor in the market.
The loss function of the firm under price $p$, given the price of the competitor $\ph$, is the negative revenue under a linear demand function:
\begin{equation*}
l(p)=\EX[-p d(p, \ph)] \coloneqq  -p(\alpha-\beta p+ \gamma \ph).
\end{equation*}
We assume $\beta>\gamma>0$, which is standard in the literature and means that the firm's demand is more sensitive to its own price than its competitor's price. 
The best response of the firm, given the competitor's price, can be easily solved as:
\begin{equation*}
p^*=\argmin_{p} l(p)=\frac{\alpha+\gamma \ph}{2 \beta}.
\end{equation*}
However, this best response requires the knowledge of $\alpha,\beta,\gamma$, which are typically unavailable to the firm. 
Next we specify how an algorithm may recommend a price based on the historical data.

%
%
%

\textbf{Algorithmic price.} 
The algorithm attempts to learn the demand function from the historical data.
However, the algorithm may not be aware of the presence of the competitor (see, e.g., \citealt{cooper2015learning}). 
Consider the gas station example in the introduction.
To provide the competitors' prices as inputs to PDI Fuel Pricing, 
the station manager needs to check the prices of nearby gas stations periodically.
Even though this is convenient, the resolution of the competitors' prices may be lower than that of the historical prices of the focal station, which have constantly been recorded in its system.
To accommodate this realistic setting with data unavailability, 
we assume that the algorithm attempts to learn a monopolistic demand function
\begin{equation}\label{equ:compe-demand-AI}
\hat d(p)=\hat\alpha-\hat\beta p.
\end{equation}
This setting of learning a monopolistic demand function under competition has also been studied in \citet{cooper2015learning} and \citet{hansen2021frontiers}.

The algorithm has access to the historical prices $p_1,\ldots,p_n$ and realized demand $d_1,\ldots,d_n$.
We assume that the demand is generated by 
\begin{equation}\label{compe-true-demand}
d_t =\alpha-\beta p_t+ \gamma p_t'+\epsilon_t,
\end{equation} 
for some independent and identically distributed (i.i.d.) noise $\epsilon_t$ and $t=1,\dots,n$.
The algorithm uses the ordinary least squares (OLS) to estimate 
$\hat{\alpha}$ and $\hat{\beta}$. 
Finally, the algorithm recommends a price maximizing the estimated revenue \eqref{compe-true-demand}, i.e., 
$\pa={\hat{\alpha}}/{2 \hat{\beta}}$.
Note that the historical demand and the algorithmic price $\pa$ depend on the unobserved competitor's prices $p_1',\ldots,p_n'$. 
To analyze $\pa$, we assume that
\begin{assumption}\label{asp:comp-pricing}
The prices $(p_n, p_n')$ are i.i.d.\ for $n=1,2,\dots$. 
Moreover, $\EX[p_n]=\EX[ p_n']=\mu$, $\Var(p_n) = \Var(p_n')=\sigma^2$ and the correlation of $(p_n, p_n')$ is $\rho\in [0,1]$.
\end{assumption}
Assumption~\ref{asp:comp-pricing} can be rather mild if we consider a symmetric duopoly and the past samples are all independent.
The condition $\rho\ge 0$ implies that the prices of competing firms are positively correlated.
The next result characterizes the asymptotic behavior of $\pa$.
\begin{lemma}\label{lem:compe-AI-price}
Suppose Assumption~\ref{asp:comp-pricing} holds.
The algorithmic price $\pa$ converges in probability to 
\begin{equation}\label{equ:compe-AI-converge}
\plim_{n\to\infty}\pa = \frac{\alpha+ \gamma \mu (1-\rho)}{2(\beta-\gamma \rho)}.
\end{equation}
\end{lemma}
To understand the algorithmic price with large data ($n\to \infty$), 
note that the symmetric Nash equilibrium price satisfies
$p_{NE}=\argmax_p\{pd(p, p_{NE})\}=\alpha/(2 \beta-\gamma)$.
If $\rho=1$, i.e., the historical prices of the firm and the competitor are perfectly correlated, then 
$\pa={\alpha}/{2(\beta-\gamma)}$
converges to the collusive price that maximizes the joint revenue of both parties.
Clearly such collusive price is higher than $p_{NE}$.
On the other hand, if $\rho=0$ and $\mu=p_{NE}$, i.e., the historical prices of the firm and the competitor are uncorrelated and centered around the Nash equilibrium,
then $\pa$ converges to $p_{NE}$.

\textbf{Human safeguard --- price matching.} 
We consider price matching, a common competitive strategy for analysts.
In particular, after receiving the price recommended by the algorithm,
the analyst may intentionally check the competitor's price.
If the competitor's price is lower than algorithmic price, 
then the analyst lowers the algorithmic price to match the competitor's price.
That is, the human-safeguarded price is  $\prin=\min\{\pa,\ph\}$.
Such an augmentation strategy is highly relevant for the human analyst: (i) 
it does not depend on the historical data or the unknown parameters, (ii) is easy to process and explain to human managers, (iii) 
it takes into account the competitive environment, and (iv) can be used to complement the algorithmic price.  
Moreover, since price matching specifies a lower bound, it also fits into the general framework in Section~\ref{sec:general-loss}.
In the next theorem, we characterize the condition under which the human augmentation improves the algorithmic price.



\begin{theorem}\label{thm:compe-pricing}
	Suppose Assumption~\ref{asp:comp-pricing} holds and the algorithmic price is given in \eqref{equ:compe-AI-converge}.
	Assume 
	$\mu \ge p_{NE}$. 
	If the competitor's price satisfies
	\begin{equation*}
	\ph \ge p_L\coloneqq \frac{\alpha \beta - 2\alpha \gamma \rho - \beta \gamma (1-\rho) \mu}{2(\beta-\rho \gamma)(\beta-\gamma)}\in (0, p_{NE}),
	\end{equation*} 
	then the revenue of the safeguarded price is higher than that of the algorithmic price, i.e., $ \prin d(\prin,p')\ge \pa d(\pa,p')$.
	In addition, if $\ph \in (p_L,\pa)$, then the revenue improvement is strictly positive.
\end{theorem}
To understand this theorem, first note that the assumption $\mu\ge p_{NE}$ is mild.
It merely states that the historical prices are a convex combination of the equilibrium price and the collusive price, since the latter is higher.
The expression for $p_L$ is complicated, 
but we can show that $p_L  \le p_{NE}$.
As long as the competitor's price is not significantly lower than the equilibrium price,
the augmentation by price matching improves the algorithmic decision. 
Price matching is particularly useful when the competitor undercuts the algorithmic price ($\ph<\pa$) while not setting a price much lower than the equilibrium price ($\ph>p_L$).

\subsection{Misspecified Algorithms}\label{sec:mis}
Consider the pricing problem in Example~\ref{exp:pricing} in a monopolistic market.
The demand function is assumed to be a general non-increasing function $f(p)$, and the unit cost of the product is $c$. 
As a result, the loss function faced by the analyst is $l(p) = -(p-c)f(p)$.
The optimal price $p^*$ satisfies $p^*=\argmin_{p} l(p)$.
Since the demand function $f$ is unknown to the analyst, 
she sets up price experiments to collect data and uses AI algorithms to learn the demand function.


\textbf{Price experiments.} 
In practice, the analyst cannot charge prices arbitrarily.
In fear of consumer backlash, the price experimentation usually takes the form of promotions, such as \$10-off coupons.
The analyst has done price experimentation at a grid of prices and observed realized demand at those price points. In particular, 
we consider the following uniform price grid between $[c,\bar p]$:
\begin{equation}\label{equ:mis-discrete-prices}
p_j=c+j\frac{\bar{p}-c}{n}, \quad j=0,1,\dots, n,
\end{equation}
where $\bar p$ represents the nominal price without any promotion.
For each price on the grid, 
we suppose $K$ noisy demand observations have been collected:
\begin{equation*}
f(p_j)+\epsilon_{jk}, \ \forall \, k={1,2,\ldots,K},
\end{equation*}
where $\epsilon_{jk}$ is an independent $\sigma$-sub-Gausssian noise. 
For example, the firm may have set the price $p_j$ for $K$ hours and recorded the hourly demand,
whose mean is $f(p_j)$ with noise $\epsilon_{jk}$.

\textbf{Algorithmic decision.}
In order to find  the optimal price $p^*$, the algorithm needs to learn the demand function $f(p)$.
However, because the price experiments are only conducted on a grid, 
the algorithm typically postulates a model for the demand function and estimate the model parameters.
One of the most common models is the linear demand function.
That is,
\begin{equation}\label{eq:mis-linear}
\hat f(p)=\hat \alpha-\hat \beta p,
\end{equation}
where $\hat\alpha$ and $\hat\beta$ guarantees that $\hat f$ is the best linear fit to the points $\{(p_j, f(p_j)+\epsilon_{jk})\}$ for $j=0,\dots, n$ and $k=1,\dots,K$ in terms of the $\ell_2$ error:
\begin{equation}\label{equ:mis-fit-MSE}
(\hat \alpha, \hat \beta)=\argmin_{\alpha,\beta} \left\{\sum_{j=0}^n \sum_{k=1}^K\left(f(p_j)+\epsilon_{jk}-\alpha+\beta p_j\right)^2\right\}.
\end{equation}
Eventually, the algorithm outputs an optimized price $p_a$ based on the estimated demand function $\hat{f}(p)$, i.e., 
$p_a=\arg\max_p (p-c)\hat f(p)= \hat \alpha/2\hat\beta+c/2$.

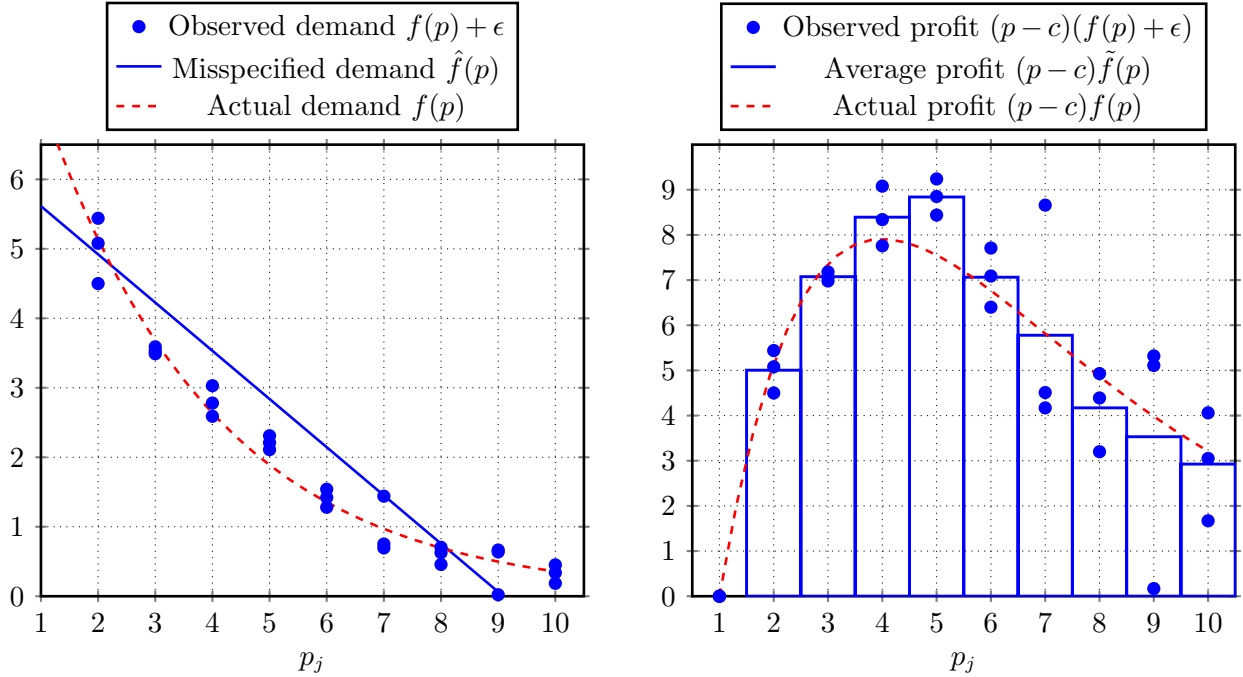
\begin{figure}[t]
	\centering
	\pgfplotsset{
		every axis/.append style={
			line width=1pt,
			tick style={line width=0.8pt}},
		scaled x ticks=false,
		tick label style={font=\normalsize},
		major grid style={dotted,color=black},
		legend style={font=\normalsize},
		tick align=center,
		width=8.8cm}
	\begin{tikzpicture}
	\begin{axis}
	[xtick={1,2,3,4,5,6,7,8,9,10},
	ytick={0,1,2,3,4,5,6},
	xmin=1, xmax=10.5,
	ymin = 0, ymax = 6.5,
	xlabel=$p_j$,
	legend style ={at={(0.5,1.03)},anchor=south
	},
	grid=major
	]
	\addplot+ [only marks, blue, mark options={fill=blue}, mark=*] table[x = xpoint, y = fy, col sep=comma] {misfigure.csv};
	\addlegendentry{Observed demand $f(p)+\epsilon$}
	\addplot [blue,domain=1:9,samples=10] {6.3087-0.694*x};
	\addlegendentry{Misspecified demand $\hat{f}(p)$}
	\addplot [red,domain=1:10,samples=30, dashed] {10*exp(-x/3)};
	\addlegendentry{Actual demand $f(p)$}
	\end{axis}
	\end{tikzpicture}
	\hfill
	\begin{tikzpicture}
	\begin{axis}[xtick={1,2,3,4,5,6,7,8,9,10},
	ytick={0,1,2,3,4,5,6,7,8,9},
	xmin=0.5, xmax=10.5,
	ymin = 0, ymax = 10,
	xlabel=$p_j$,
	legend style ={at={(0.5,1.03)},anchor=south},
	grid=major,
	extra x tick style={
		x tick label as interval=false
	}
	]
	\addplot+ [only marks, blue, mark options={fill=blue}, mark=*] table[x = xpoint, y = xfy, col sep=comma] {misfigure.csv};
	\addlegendentry{Observed profit $(p-c) (f(p)+\epsilon)$}
	\addplot[line width=1.1pt,ybar interval,
	draw=blue
	] 
	coordinates
	{
		(1.5,5.0045) (2.5,7.0755) (3.5,8.3929) (4.5,8.84) (5.5,7.0635) (6.5,5.7788) (7.5,4.1711) (8.5,3.5318) (9.5,2.926) (10.5,2.926)
	};
	\addlegendentry{Average profit $(p-c) \tilde f(p)$}
	\addplot [red,domain=1:10,samples=30, dashed] {10*(x-1)*exp(-x/3)};
	\addlegendentry{Actual profit $(p-c)f(p)$}
	\end{axis}
	\end{tikzpicture}
	\caption{The left figure shows the misspecified linear model by the algorithm. The actual demand function is $f(p)=10 \exp(-p/3)$ with $c=1$, $\bar{p}=10$, $n=10$, and $K=3$. The OLS estimator is $\hat{\alpha}=6.309$, $\hat{\beta}=-0.694$. The right figure shows the human analyst identifying the price that earns the highest empirical profit.}
	\label{fig:mis-linear}
\end{figure}

Although linear demand models have been shown in \citet{cohen2021simple} to perform well when the model is misspecified,
in our setup, we do not claim that the linear model is necessarily the best algorithmic choice in this scenario.
In fact, given the data points $\{(p_j, f(p_j)+\epsilon_{jk})\}$, there may be other choices such as a linear interpolation that can fit the demand function better.
Our main goal is to demonstrate a salient feature of a wide range of algorithms based on parametric statistical models---the chosen model may be misspecified and hence the resulting algorithm may be built on a shaky foundation. 
That is, the relationship between price and demand may not be  accurately captured by the postulated class of models. 
In this case, even with sufficient data, the misspecification cannot be fully remedied.
As we shall see in the left of Figure~\ref{fig:mis-linear}, such a  misspecification can be clearly pronounced for the linear model.

\textbf{Human knowledge.} 
How can human knowledge help with the misspecified algorithm?
Due to the limitation of the human brain, the human analyst typically does not form a model to process the historical data, and it would be impossible to judge whether the algorithmic price suffers from misspecification.
We consider a rather natural and straightforward approach:
since the analyst observes the noisy demand on the price grid, 
it first uses the average to form an estimate of the demand at each price as 
\begin{equation}
\tilde{f}(p_j) \coloneqq f(p_j)+\frac{1}{K} \sum_{k=1}^K \epsilon_{jk}.
\end{equation}
Using this estimate,
the optimal price on the grid that generates the highest empirical profit $(p_j-c) \tilde f(p_j)$ can be easily calculated. 
Suppose $j^*$ is the index of one of the optimal prices on the grid:
\begin{equation*}
    (p_{j^*}-c) \tilde f(p_{j^*})\ge (p_{j}-c) \tilde f(p_{j})\quad \forall j=0,\dots, n.
\end{equation*}
Taking the demand function in Figure~\ref{fig:mis-linear} as an example, the human analyst observes the noisy demand on the price grid $\{1,2,\ldots,10\}$. Then, she chooses the price point $p_{j^*}=5$ that gives the highest empirical profit. 
In this example, the chosen price $p_{j^*}=5$ does not equal to the true optimal price $p^*=4$, but the neighborhood $[p_{j^*-1},p_{j^*+1}]=[4,6]$ includes $p^*$.
It is easy to see the complementary effects of human knowledge and the algorithm in this example.
The optimal price from the human knowledge is empirically validated without any statistical model. 
However, while the algorithm may suffer from misspecification, 
it has two strengths unmatched by the human analyst.
First, the algorithm aggregates all $(n+1)K$ demand observations while the human analyst takes the average of $K$ demand observations locally. 
It is well-known that more samples improve the statistical prediction power.
Second, the human analyst does not attempt to specify a model to extrapolate the demand function. 
As a result, only the prices on the grid
can be selected, and a discretization error is always born by the price picked by the analyst.
For example, if the prices $\{\bar p-20, \bar p-10, \bar p\}$ have been experimented, i.e., two types of promotions, $\$20$ off and $\$10$ off, in addition to the nominal price $\bar p$, have been offered in the past,
then $p_{j^*}$ can be suboptimal if the actual optimal price is $\bar p-15$. 
In this case, the algorithm learns a model that interpolates the price gaps on the grid and remedies the discretization error.
The following result characterizes the 
performances of the two approaches, which allow us to further understand the benefit of human augmentation to the algorithm.
\begin{proposition}\label{prop:mis-finite-human}
Assume that the loss function $l(p)$ is strongly convex with parameter  $\lambda$ (or equivalently, the profit function is $\lambda$-concave), i.e., 
\begin{equation*}
l(p) \le l(p')-l'(p') (p-p')-\frac{\lambda}{2} (p-p')^2, \quad \forall \, p,p' \in [c,\bar{p}].
\end{equation*}
We then have
\begin{enumerate}
    \item [(\romannumeral1)][Algorithmic decision] Let $\pas$ denote the optimal price for the misspecified linear demand. Given $n$ and $K$, we have
	\begin{equation}\label{equ:mis-sample-AI}
	\PR(|\pa-\pas| \ge \delta) \le 4 \exp(-b nK), 
	\end{equation}
	where $b$ is a constant independent of $n$ and $K$.
	\item [(\romannumeral2)][Human knowledge] The probability of the true optimal price not falling into the neighborhood of human's estimated price $p_{j^*}$ satisfies
	\begin{equation}\label{equ:mis-sample-human}
    \PR\left(p^* \notin [p_{j^*-1},p_{j^*+1}] \right) \le  2(n+1) \exp\left(-\frac{K \lambda^2 (\bar{p}-c)^4}{32 \sigma^2 \bar{p}^2 n^4}\right).
\end{equation}
\end{enumerate}
\end{proposition}
Note that for large samples ($K\to\infty$ or $n\to\infty$), the algorithmic decision
does not converge to the true optimal price $p^*$.
Instead, it converges to the optimal price for the misspecified linear demand.
For the human knowledge, without misspecification,
the price neighborhood $[p_{j^*-1},p_{j^*+1}]$ on the grid containing $p_{j^*}$ will eventually include the true optimal price as $K\to\infty$.
However, compared to the algorithmic decision, the human's error probability can be significantly inflated, which even increases in the number $n$ of price points, reflecting a lack of data efficiency, while the algorithm's error probability decreases in $n$.

\textbf{Augmentation by safeguarding.} 
Because of the weaknesses of the finite-sample results in Proposition~\ref{prop:mis-finite-human} (i.e., the convergence to the wrong target in \eqref{equ:mis-sample-AI} by the algorithm and inefficient sample use in \eqref{equ:mis-sample-human} by the human),
the analyst may decide to integrate both approaches.
Based on $p_{j^*}$, the human analyst imposes a guardrail $[p_{j^*-1}, p_{j^*+1}]$ as in Proposition~\ref{prop:loss-two-side}. 
In other words, the human analyst uses the two neighboring prices of  $p_{j^*}$ on the grid to form an interval to regulate the algorithmic output. 
As a result, the safeguarded algorithmic price is 
\begin{equation*}
    \prin=\max\{\min\{\pa,p_{j^*+1}\},p_{j^*-1}\}. 
\end{equation*}
The following result characterizes the condition under which such an  augmentation is beneficial.
\begin{theorem}\label{thm:misspec}
Assume the profit function $(p-c)f(p)$ is unimodal. 
The augmentation improves the algorithmic price, i.e.,
$(\prin-c)f(\prin) \ge (\pa-c)f(\pa)$, if the true optimal price 
$p^* \in [p_{j^*-1},p_{j^*+1}]$.
In particular, the latter condition always holds when $K\to\infty$.
\end{theorem}
Theorem~\ref{thm:misspec} requires the profit function to be unimodal, which is satisfied by most demand functions $f(\cdot)$ (see, e.g.,  \citealt{ziya2004relationships}). 
As a result, the regime that sees the most benefit of human augmentation is when $n$ is fixed but $K\to\infty$, i.e., the price experimentation is conducted on a few prices for an extended period.
This is arguably a common scenario in retailing, due to the infeasibility of frequent price changes.
In this case, augmenting the algorithmic price using the bounds distilled from the human knowledge,
the augmented price $\hat p$ enjoys the benefits of both worlds.
Intuitively, when $p^*$ falls in the interval $[p_{j^*-1}, p_{j^*+1}]$
and the algorithmic price is outside the interval, the human safeguard always 
pulls the algorithmic price toward the actual optimal price and
improves the algorithmic recommendation due to the unimodality of the profit function.
On the other hand, when the algorithmic price falls into the same interval, indicating the discretization error may exceed the misspecification error (imagine $[p_{j^*-1}, p_{j^*+1}]$ being a wide interval), the guardrail does not take effect and the analyst follows the algorithmic decision.
This result confirms the complementary effects of algorithms and human knowledge, in particular, the robustness of simple heuristics against model misspecification.

We next study two commonly used demand function forms of $f(\cdot)$  and characterize the conditions under which the algorithmic price falls outside the interval $[p_{j^*-1}, p_{j^*+1}]$, i.e., when it is strictly improved by the human augmentation.
We consider $K\to\infty$ in both examples.
\begin{example}[Isoelastic demand]
	Consider the demand function $f(p)=bp^{-a}$ where $a>1$ and $b>0$. 
	It can be shown that the profit function is unimodal, and the optimal price is
    $p^*=\frac{ac}{a-1}$. 
    We can show that when the nominal price 
	\begin{equation}\label{equ:mis-poly}
	\bar{p} > \dfrac{\frac{a}{a-1}-\frac{1}{2}-\frac{2}{n}}{\frac{1}{3}-\frac{2}{n}} c,
	\end{equation}
	the algorithmic price $\pa$ is outside the interval $[p_{j^*-1}, p_{j^*+1}]$ and thus the human  augmentation strictly improves the algorithm.
	For example, if $a=2$, $n=10$, then \eqref{equ:mis-poly} is equivalent to $\bar{p}>2.25c$, i.e., when the nominal price is more than 125\% higher than the production cost.
\end{example}

\begin{example}[Exponential demand]
	Consider the demand function $f(p)=b e^{-ap}$ where $a>0$. 
	It can be shown that the optimal price is $p^*=1/a+c$.
	We can show that when
	\begin{equation}\label{equ:mis-exp}
	\bar{p} > \dfrac{\frac{1}{a}+(\frac{1}{2}-\frac{2}{n})c}{\frac{1}{3}-\frac{2}{n}},
	\end{equation}
	the human augmentation strictly improves the algorithm.
	For example, if $a=2$, $n=10$, then \eqref{equ:mis-exp} is translated to $\bar{p}>3.75+2.25c$.
\end{example}
From both examples, we can see that the misspecification error by the algorithm grows larger relative to the discretization error by the guardrail and hence human's augmentation becomes more beneficial, when $\bar p $ is much larger than $c$, i.e., the interval for the price experiments is wider. 

\subsection{Data Contamination}\label{sec:conta-LR}
In this section, we consider the case when the data can be contaminated, due to outliers, reporting errors, etc.
While the possibly contaminated data is fed into algorithms, the error in data 
also propagates to the output decision.
For this reason, \cite{fogliato2022case} advocate humans-in-the-loop, to mitigate the data contamination.
Human analysts are less susceptible to data contamination because the human brain cannot process large data sets, which turns out to be a blessing rather than a curse in this case as it makes the human knowledge robust to minor data contamination. 
Next we provide a formal analysis of human augmentation to the algorithms for this use case. 

The application we consider is the linear regression problem in Example \ref{example:MSE-LR}. 
Without contamination,
the historical data $\{(X_i, W_i)\}_{i=1}^n$ is generated by
$X_i=W_i^\top \beta+ \epsilon_i$
for some unknown coefficients $\beta$ and noise $\epsilon_i$.
We consider two contamination mechanisms that affect a fraction of samples: 
contamination in response \citep{bhatia2017consistent} and covariates \citep{mcwilliams2014fast,loh2011high}. 
Before we provide the formal introduction to the mechanisms,  we first
explain how the algorithm and human knowledge play a role in the process.

Not able to tell whether the data is contaminated,
the algorithm simply applies the OLS estimator to the data\footnote{We acknowledge that there are statistical tests to identify outliers and robust estimators to mitigate data contamination. We do not consider them in the model because they usually require some information about the contamination such as whether the data is contaminated or the contamination mechanism, while in practice, the algorithm is agnostic to such knowledge.}, as stated in Example~\ref{example:MSE-LR}.
For a new covariate $W$, we denote the prediction from the OLS estimator as $X_a(W)$.
For the human analyst,
we consider a generic two-sided bounded range $[\xhl,\xhu]$. 
As shown in Section~\ref{sec:general-loss},
the corresponding safeguarded decision is  $\xin(W)=\max\{\min\{\xa(W),\xhu\},\xhl\}$ for the two-sided guardrail. 
Also recall that the loss function is $l(x,w)=(x-w^\top \beta)^2$.



\subsubsection{Contamination in Response}\label{sec:conta-resp}
We first consider the contamination in the response of the samples.
In particular, the observed response $X_i$ is not generated from $W_i^\top \beta+ \epsilon_i$, 
but 
\begin{equation}\label{eq:contam-response}
  X_i=W_i^\top \beta +B_i+ \epsilon_i  
\end{equation}
for some random variable $B_i$.
Here $B_i$ controls for the degree of contamination: with a high probability, it is zero and the sample is not contaminated.
When $B_i\neq 0$, the response of the sample, $X_i$, deviates from the uncontaminated observation $W_i^\top \beta + \epsilon_i$.
Note that $B$ is not to be confused with $\epsilon$, which has a zero mean.  We assume $\EX[B]$ to be nonzero which means that the contamination has a systematic influence on the estimation. 

Contamination in the response is studied in the computer science community, see, e.g.,  \cite{wright2008robust} and \cite{nguyen2012robust} for applications to image recognition.
For management applications, this type of contamination 
may occur due to various reasons:
the historical response $X_i$ may be subject to reporting errors, or may be censored in certain periods and some ad hoc imputation methods are used so that the missing data are replaced by their estimated values.
Our contamination model can capture both cases. 

Not surprisingly, when the historical data is contaminated, the algorithmic output that uses OLS is biased.
More precisely, given the i.i.d. historical samples $\{(X_i, W_i)\}_{i=1}^N$ that are generated by \eqref{eq:contam-response} and a new covariate $w$, 
suppose $\xa(w)$ is the OLS estimator applied to $w$. 
We have: 
\begin{lemma}\label{lem:conta-res}
	Assume the covariance matrix $\Sigma=\EX[W W^\top]$ is positive definite. Then the OLS predictor $\xa(w)=w^\top\hat{\beta}$ converges to $w^\top \beta +\EX[B]$ in probability for any $w$ as $N\to\infty$.
	Therefore, $l(\xa(w),w)=\EX[B]^2$.
\end{lemma}
In other words, even with a sufficiently large dataset, 
the bias caused by the contamination persists.
To analyze how the human augmentation may improve the algorithmic result, we consider a simple form of contamination.
Let $B=b>0$ with probability $p$ and $B=0$ with probability $1-p$.
Therefore, the contamination always leads to upward bias (the downward bias can be formulated similarly), and the parameter $b$ represents the magnitude of the contamination and $p$ represents its propensity.
To adjust for the upward bias, the human analyst can impose an upper bound $\xhu$ and cap the output of the algorithm and lead to the safeguarded prediction as  $\xin(W)=\min\{\xa(W),\xhu\}$.
However, because $\xhu$ is not data-driven, the safeguard runs the risk of overcorrection.
The following proposition provides such a condition.


\begin{proposition}\label{prop:conta-response}
	Assume the domain of $W$ is a closed and bounded set $\mathcal{W} \in \mathbb{R}^d$, and we take $N\to\infty$. If $\xhu$ satisfies
	\begin{equation}\label{equ:conta-LR-response}
	\xhu \ge \max_{w\in \mathcal{W}}\{w^\top \beta\}-p b,
	\end{equation}
	then, for all $w\in \mathcal {W}$, we have $l(\xin(w),w ) \le l(\xa(w),w ) $. 
\end{proposition}
To interpret the result, on the one hand, in the extreme case of $\xhu \ge\max_{w \in \mathcal{W}} \{w^\top \beta\}+pb $, we always have $\xhu \ge \xa$ and $\xin=\xa$. The bound of $\xhu$ is too conservative, and the safeguard provides no benefit. 
On the other hand, 
one can show that the loss function for the 
algorithmic prediction is simply
$l(\xa(w),w) = p^2 b^2$ due to the contamination.
If $\xhu < \max_{w \in \mathcal{W}} \{w^\top \beta\}-pb$, 
then there exists $w$ such that $l(\xin({w}),{w}) > p^2 b^2$ because the upper bound imposed by the human analyst is too aggressive and outweighs the bias introduced by the contamination.
Clearly, condition~\eqref{equ:conta-LR-response}
is easier to satisfy when the contamination gets more severe due to an increased value of $p$ or $b$.

When the contamination can lead to a bias of either direction, i.e., it is possible that $B>0$ or $B<0$,
it is safer for the human analyst to set up both bounds $\xhl$ and $\xhu$.
That is, once the algorithmic prediction $\xa(w)$ is given, 
the safeguarded decision is $\xin(w)=\max\{\min\{\xa(w),\xhu\},\xhl\}$.
Generalizing Proposition~\ref{prop:conta-response}, we have: 
\begin{theorem}\label{thm:conta-respon}
	Suppose the domain of $W$ is a closed and bounded set $\mathcal{W} \in \mathbb{R}^d$, and we take $N\to\infty$. If the lower and upper bounds $ \xhl,\xhu$ satisfy
	\begin{equation}\label{equ:conta-LR-response-two-side}
	 \xhu \ge \max_{w\in \mathcal{W}}\{w^\top \beta\}- \big|\EX[B]\big|, \ \  \xhl \le \min_{w\in \mathcal{W}}\{w^\top \beta\}+ \big|\EX[B]\big|,
	\end{equation}
	then for all $w\in \mathcal {W}$, we have $ l(\xin(w),w ) \le l(\xa(w),w )$ and $\EX[l(\xin(W),W)] \le  \EX[l(\xa(W),W)]$.
\end{theorem}
Note that when the absolute bias ($\big|\EX[B]\big|$) is large,
the human augmentation of imposing upper and lower bounds tends to be helpful, regardless of the sign of the bias.
For example, even when $B>0$, i.e., the bias is always upward, 
Theorem \ref{thm:conta-respon} states that  
imposing a lower bound $\xhl$ is more likely to be beneficial when the bias is large because    \eqref{equ:conta-LR-response-two-side} is more likely to be satisfied. 
To see the intuition, as the contamination becomes more severe, the algorithmic decision is subject to a larger bias. Hence, it is easier for the human augmentation to outperform the raw algorithmic decision.

%
%

\subsubsection{Contamination in Covariates}
In this section, we consider the covariates in the data, $W_i$, which can be contaminated.
This type of contamination is sometimes referred to as the error-in-variable  \citep{loh2011high}, which occurs in voting, surveys, and sensor networks.
In business applications, 
there may exist measurement errors in the historical samples of the covariate.
For example, when a firm is running a survey to learn consumer sentiment, 
the design of the survey may lead to biased measurement of the quantity of interest.

We consider the following contamination model: the observed covariate is generated by $W_i=Z_i+U_i$,  
where $Z_i$ is the actual covariate, and $U_i\in \mathbb{R}^d$ is an error that contaminates the observation, independent of $Z_i$.
The response is generated from $X_i=Z_i^\top \beta+\epsilon_i$.
For a new covariate $W_0$, the algorithm outputs $\xa(W_0)$ using the OLS estimator from the data $\{(X_i,W_i\}_{i=1}^\infty$. (We consider infinite samples in this  analysis.)

What differentiates the contamination in covariates from Section~\ref{sec:conta-resp} is that even the new covariate $W_0$ itself may be contaminated. 
Therefore, the human safeguard can serve for two purposes:
it helps to control the contamination in the training data and curtail the potential error in the new covariate based on which the prediction is given.
We impose the following technical assumption.
\begin{assumption}\label{ass:conta-cov-design}
	The matrix $\Sigma_1 \coloneqq \EX[ZZ^\top]$ is positive definite and $\Sigma_2 \coloneqq \EX[U U^\top]$ is positive semi-definite.
\end{assumption}
Next, we show that the contamination usually leads to an inconsistent OLS estimator.
\begin{lemma}\label{lem:conta-cov}
	 Suppose Assumption \ref{ass:conta-cov-design} holds. 
	 The OLS estimator $\hat{\beta}$ for \eqref{eq:contam-response} converges to $\left(\mathcal{I}-(\Sigma_1+ \Sigma_2)^{-1} \Sigma_2\right) \beta$ in probability. Furthermore, $\hat{\beta}$ converges to $\beta$ in probability if and only if  $\Sigma_2 \beta=\bm{0}$.
\end{lemma}
From Lemma~\ref{lem:conta-cov}, we know that the OLS estimator $\hat{\beta}$ does not converge to the true parameter $\beta$ unless $\beta$ is in the null space of $\Sigma_2$. 
As a result, the predicted response $\xa(w)=w^\top \hat{\beta}$ given 
$w$ is usually biased. 
Note that in this case the bias can be translated to the contamination in response as in Section~\ref{sec:conta-resp}, and the conditions in Theorem~\ref{thm:conta-respon} can be similarly applied.
In this section, we instead focus on a different angle:
even when $\Sigma_2 \beta=\bm{0}$ holds and $\hat{\beta}$ converges to $\beta$, the algorithm is still not bias-free.
This is because the new covariate $W_0$ may be contaminated.
The actual prediction should be $Z_0^\top \beta = (W_0-U_0)^\top\beta$, while under contamination, even with $\hat\beta = \beta$, the prediction is $W_0^\top\beta$.

We consider the two-sided guardrail $[\xhu,\xhl]$ for the human augmentation.
Note that in this case, the loss functions for the algorithmic and safeguarded outcomes are $l(\xa(W), Z)$ and $l(\xin(W),Z)$, respectively, 
because the loss only depends on the actual covariate $Z$, not the observed but potentially contaminated $W$.
We impose the following technical assumption.
\begin{assumption}\label{ass:conta-cov-VW}
	Assume the domain of $Z$ is a closed and bounded set $\mathcal{Z} \in \mathbb{R}^d$. Assume $\Sigma_2 \beta=\bm{0}$ and there exist constants $b,p \in (0,0.5)$ such that
	\begin{equation}\label{equ:LR-conta-cov-v}
	\PR\left(U^\top\beta \ge b\right) \ge p, \  \PR\left(U^\top\beta \le -b\right) \ge p.
	\end{equation}
\end{assumption}
The compactness of the covariate $Z$ is similar to the assumption in Theorem~\ref{thm:conta-respon}. 
Equation~\eqref{equ:LR-conta-cov-v} states that the contamination $U^\top \beta$ is not concentrated at zero, which would make the application of the two-sided range more likely to be  beneficial. 
Next we state our main result.
\begin{theorem}\label{thm:conta-covariates}
	Suppose Assumptions  \ref{ass:conta-cov-design} and \ref{ass:conta-cov-VW} hold. If the upper and lower bounds $\xhu,\xhl$ satisfy
	\begin{equation}\label{equ:conta-cov-discrete}
	\xhl \le \min_{z\in \mathcal{Z}}\{z^\top \beta\}+\sqrt{\frac{p}{1-p}}b, \ \  \xhu \ge \max_{z\in \mathcal{Z}}\{z^\top \beta\}-\sqrt{\frac{p }{1-p}}b,
	\end{equation} 
	then we have $\EX[l(\xin(W),Z)] \le  \EX[l(\xa(W),Z)]$. 
\end{theorem}
%
%

Comparing to Theorem~\ref{thm:conta-respon}, 
it is worth pointing out that although the contamination mechanisms differ, the conditions for beneficial human augmentation are surprisingly similar.
In particular, when $b$ or $p$ is larger, i.e., the magnitude of the contamination increases,
there is more room for human augmentation to be helpful (as conditions (\ref{equ:conta-cov-discrete}) are more likely to hold).

\section{Conclusion}\label{sec:conclusion}
Motivated by a consulting project on retail fuel pricing, we propose a framework to study the human-AI interaction in which an algorithm first recommends a decision to the human analyst, then the analyst can augment it based on domain knowledge and experience. 
As far as we know, this is the first study to investigate this type of interaction.
With the framework, we investigate when human knowledge adds value to algorithmic decision-making. 
We demonstrate three common and practical situations in which human knowledge may play a critical role in harnessing and correcting algorithmic decisions, even with large data.

We conclude by discussing potential future directions. 
First, we may consider more sophisticated yet realistic human augmentation.
For example, AI is known to suffer from out-of-distribution issues when the algorithmic decision learned from the training data does not provide much value, and for these instances, human knowledge is particularly useful in correcting. 
It is desirable to extend our framework and incorporate simple rules to identify such instances and override the algorithmic decision. 
Second, another important reason for humans to intervene is the consideration for fairness or ethical issues associated with the algorithmic decision. 
It is a fruitful direction to extend our framework by incorporating these considerations as rules of thumb to guardrail algorithmic outputs. 
Lastly, in some applications, the adoption choice between the algorithm and the human knowledge needs to be made before the computation of algorithmic decisions because it may incur significant waiting if human correction is conducted after observing the algorithmic decisions.
In this case, the human analyst needs to design and commit to a simple rule based on the observed covariate. 
Our framework may be extended to study this kind of human-AI interaction.

\bibliographystyle{informs2014}
\bibliography{myrefs_AI_DDM} 

\newpage
\setcounter{page}{1}
\setcounter{table}{0}
\setcounter{section}{0}
\numberwithin{table}{section}
\begin{center}
	{\Large Online Appendix to\\
		``Algorithmic Decision-Making Augmented by Human Knowledge''}
\end{center}
\renewcommand\thesection{\Alph{section}}

\section{Proofs in Section \ref{sec:general-loss}}
\textbf{Proof of Proposition \ref{prop:loss-one-side-up}.}
	To prove (\romannumeral1), we first write down the expression for $\EX[l(\xin)]$,
	\begin{align*}
	\EX[l(\xin)] 
	&= \int_{-\infty}^{+\infty} \int_{-\infty}^{+\infty} l \left(\min(\sxa,\sxh)\right) f(\sxa,\sxh) \dif \sxa \dif \sxh \\
	&=\int_{-\infty}^{+\infty} \left(\int_{-\infty}^{\sxh} l(\sxa) f(\sxa,\sxh) \dif \sxa + \int_{\sxh}^{\infty} l(\sxh) f(\sxa,\sxh)\dif \sxa \right) \dif \sxh,
	\end{align*}
	where $f(\sxa,\sxh)$ denotes the joint probability density function of $\xa, \xh$. The last equality follows from separating the integral in $\xa$ by $\xh \le \xa$ and $\xh > \xa$.
	Then, we have
	\begin{align}\label{equ:MSE-minus}
	\EX[l(\xa)]-\EX[l(\xin)] &=\int_{-\infty}^{\infty} \int_{\sxh}^{\infty} \left(l(\sxa) -l(\sxh)\right) f(\sxa,\sxh) \dif \sxa \dif \sxh\\
	&=\EX[\left(l(\xa)-l(\xh)\right) \mathbb{I} (\xh \leq \xa)],\notag
	\end{align}
	which completes the proof of (\romannumeral1).
	
	To prove (\romannumeral2), we separate the integral in \eqref{equ:MSE-minus} by $\xh\leq \xx$ and $\xh > \xx$:
	\begin{align}
	\EX[l(\xa)]-\EX[l(\xin)]
	&=\int_{-\infty}^{\xx} \int_{\sxh}^{\infty} \left(l(\sxa) -l(\sxh)\right) f(\sxa,\sxh) \dif \sxa \dif \sxh \notag\\ 
	& \quad+\int_{\xx}^{\infty} \int_{\sxh}^{\infty} \left(l(\sxa) -l(\sxh)\right) f(\sxa,\sxh) \dif \sxa \dif \sxh. \label{equ:loss-two-intergral}
	\end{align}
	For the first term in \eqref{equ:loss-two-intergral}, we have
	\begin{align}
	&\int_{-\infty}^{\xx} \int_{\sxh}^{\infty} \left(l(\sxa) -l(\sxh)\right) f(\sxa,\sxh) \dif \sxa \dif \sxh \notag\\
	&\overset{(a)}{=}\int_{-\infty}^{\infty} \int_{-\infty}^{\min\{\xx,\sxa\}} \left(l(\sxa) -l(\sxh)\right) f(\sxa,\sxh) \dif \sxh \dif \sxa \notag\\
	&\overset{(b)}{=}\int_{-\infty}^{\xx} \int_{-\infty}^{\sxa} \left(l(\sxa) -l(\sxh)\right) f(\sxa,\sxh) \dif \sxh \dif \sxa+\int_{\xx}^{\infty} \int_{-\infty}^{\xx} \left(l(\sxa) -l(\sxh)\right) f(\sxa,\sxh) \dif \sxh \dif \sxa \notag \\
	&=\EX[\left(l(\xa)-l(\xh)\right) \mathbb{I} (\xa \le \xx, \xh \leq \xa)]+ \EX[\left(l(\xa)-l(\xh)\right) \mathbb{I} (\xa > \xx, \xh \leq \xx )] \notag \\
	&=\EX[\left(l(\xa)-l(\xh)\right) \mathbb{I} (\xh \leq \xa \le \xx)]+ \EX[\left(l(\xa)-l(\xh)\right) \mathbb{I} (\xh \leq \xx < \xa)]. \label{equ:loss-one-low-x},
	\end{align}
	where $(a)$ follows from changing the order of integration, and $(b)$ follows from separating the integral by $\xa \le \xx$ and $\xa 
	> \xx$.
	
	For the second term in \eqref{equ:loss-two-intergral}, we have
	\begin{align}
	&\int_{\xx}^{\infty} \int_{\sxh}^{\infty} \left(l(\sxa) -l(\sxh)\right) f(\sxa,\sxh) \dif \sxa \dif \sxh \notag \\
	&\overset{(a)}{=}\int_{\xx}^{\infty} \int_{\xx}^{\sxa} \left(l(\sxa) -l(\sxh)\right) f(\sxa,\sxh) \dif \sxh \dif \sxa \notag \\
	&=\EX[(l(\xa)-l(\xh)) \mathbb{I} (\xa > \xx, \xx < \xh \leq \xa)]\notag \\
	&=\EX[(l(\xa)-l(\xh)) \mathbb{I} (\xx < \xh \leq \xa)], \label{equ:loss-one-high-x}
	\end{align}
	where $(a)$ follows from changing the order of integration. 
	Plugging \eqref{equ:loss-one-low-x} and \eqref{equ:loss-one-high-x} into \eqref{equ:loss-two-intergral}, we have  
	\begin{align}
	\eqref{equ:loss-two-intergral}
	&=\EX[\left(l(\xa)-l(\xh)\right) \mathbb{I} (\xh \leq \xa \le \xx)]+ \EX[\left(l(\xa)-l(\xh)\right) \mathbb{I} (\xh \leq \xx < \xa)] \notag\\
	&\quad+\EX[(l(\xa)-l(\xh)) \mathbb{I} (\xx < \xh \leq \xa)] \label{equ:MSE-combine-original}\\
	&=\EX[\left(l(\xa)-l(\xx)\right) \mathbb{I} (\xh \leq \xa \le \xx)]+\EX[\left(l(\xx)-l(\xh)\right) \mathbb{I} (\xh \leq \xa \le \xx)]\notag\\
	&\quad+ \EX[\left(l(\xa)-l(\xx)\right) \mathbb{I} (\xh \leq \xx < \xa)]+\EX[\left(l(\xx)-l(\xh)\right) \mathbb{I} (\xh \leq \xx < \xa)]\notag\\
	&\quad+\EX[(l(\xa)-l(\xh)) \mathbb{I} (\xx < \xh \leq \xa)], \label{equ:MSE-combine}
	\end{align}
	where in the last equality, we separate $l(\xa)-l(\xh)$ into $l(\xa)-l(\xx)$ and $l(\xx)-l(\xh)$.
	By Assumption \ref{ass:general-loss} (\romannumeral2), we have $l(\cdot) \ge l(\xx)$ and $l(x_1) \ge l(x_2)$ for $x_1 \ge x_2 \ge \xx$. Thus, we have
	\begin{align}
	&\EX[\left(l(\xa)-l(\xx)\right) \mathbb{I} (\xh \leq \xa \le \xx)] \geq 0, \label{equ:loss-left-bottom} \\
	&\EX[\left(l(\xa)-l(\xh)\right) \mathbb{I} (\xx < \xh \leq \xa)] \ge 0. \label{equ:loss-right-top}
	\end{align}
	Furthermore, since $l(\xx) \le l(\cdot)$ and $\mathbb{I} (\xh \leq \xa \le \xx) \le \mathbb{I} (\xa \leq \xx, \xh \leq \xx)$, we have
	\begin{align}
		\EX[\left(l(\xx)-l(\xh)\right) \mathbb{I} (\xh \leq \xa \le \xx)] \geq \EX[\left(l(\xx)-l(\xh)\right) \mathbb{I} (\xa \leq \xx, \xh \leq \xx)], \label{equ:loss-left-bottom-z}
	\end{align}
	By \eqref{equ:loss-left-bottom}, \eqref{equ:loss-right-top}, the first and last term in \eqref{equ:MSE-combine} can be lower-bounded by zero, and the second term in \eqref{equ:MSE-combine} has the lower bound in \eqref{equ:loss-left-bottom-z}.
	Thus, plugging \eqref{equ:loss-left-bottom}, \eqref{equ:loss-left-bottom-z} and \eqref{equ:loss-right-top} into \eqref{equ:MSE-combine}, we have 
	\begin{align}
	\eqref{equ:MSE-combine} 
	& \ge \EX[\left(l(\xx)-l(\xh)\right) \mathbb{I} (\xa \leq \xx, \xh \leq \xx)]+\EX[\left(l(\xa)-l(\xx)\right) \mathbb{I} (\xh \leq \xx < \xa)] \notag \\
	&\quad+\EX[\left(l(\xx)-l(\xh)\right) \mathbb{I} (\xh \leq \xx < \xa)]\notag \\
	&\overset{(a)}{=} \EX[(l(\xa)-l(\xx)) \mathbb{I} (\xh \leq \xx < \xa)]+ \EX[(l(\xx)-l(\xh)) \mathbb{I} (\xh \leq \xx)] \notag \\
	&=\EX[(l(\xa)-l(\xx)) \mathbb{I} (\xh \leq \xx < \xa)] -\EX[(l(\xh)-l(\xx)) \mathbb{I} (\xh \leq \xx)] \notag \\
	&\overset{(b)}{=}\EX[l(\xa) \mathbb{I} (\xh \leq \xx < \xa)]- \EX[l(\xh) \mathbb{I} (\xh \leq \xx)]+l(\xx) \PR(\xa \le \xx, \xh \le \xx) \notag\\
	& \ge \EX[l(\xa) \mathbb{I} (\xh \leq \xx < \xa)]- \EX[l(\xh) \mathbb{I} (\xh \leq \xx)], \label{equ:MSE-combine-ge-zero}
	\end{align}
	where $(a)$ holds by $\mathbb{I} (\xa \leq \xx, \xh \leq \xx)+\mathbb{I} (\xh \leq \xx < \xa)=\mathbb{I} (\xh \leq \xx)$, $(b)$ holds by $\mathbb{I} (\xh \leq \xx)-\mathbb{I} (\xh \leq \xx < \xa)=\mathbb{I}(\xa \le \xx, \xh \le \xx)$ and $(c)$ follows from $l(\xx) \ge 0$ due to Assumption \ref{ass:general-loss} (\romannumeral1).
	Thus, if $\eqref{equ:MSE-combine-ge-zero} \geq 0$, we have $\eqref{equ:loss-two-intergral} \ge 0$ and $\EX[l(\xin)] \le \EX[l(\xa)]$, which completes the proof of (\romannumeral2).
	
	To prove (\romannumeral3), note that 
	\begin{align}
	\eqref{equ:MSE-combine-original}
	&=\EX[\left(l(\xa)-l(\xh)\right) \mathbb{I} (\xh \leq \xa \le \xx)]+ \EX[\left(l(\xa)-l(\xh)\right) \mathbb{I} (\xh \leq \xx < \xa)] \notag\\
	&\quad+\EX[(l(\xa)-l(\xx)) \mathbb{I} (\xx < \xh \leq \xa)]+\EX[(l(\xx)-l(\xh)) \mathbb{I} (\xx < \xh \leq \xa)] \label{equ:MSE-combine-upper}
	\end{align}
	By Assumption \ref{ass:general-loss} (\romannumeral2), we have
	\begin{align}
	&\EX[(l(\xa)-l(\xh)) \mathbb{I} (\xh \leq \xa \le \xx)] \leq 0, \label{equ:loss-right-top-up}\\
	&\EX[(l(\xx)-l(\xh)) \mathbb{I} (\xx < \xh \leq \xa)] \leq 0. \label{equ:loss-right-bottom-up}
	\end{align}
	Plugging \eqref{equ:loss-right-top-up} and \eqref{equ:loss-right-bottom-up} into \eqref{equ:MSE-combine-upper}, the first and last term in \eqref{equ:MSE-combine-upper} are upper bounded by zero. And we have 
	\begin{align}
	\eqref{equ:MSE-combine} 
	&\leq \EX[(l(\xa)-l(\xh)) \mathbb{I} (\xh \leq \xx < \xa)]+\EX[(l(\xa)-l(\xx)) \mathbb{I} (\xx < \xh \leq \xa)] \notag\\
	& \overset{(a)}{=} \EX[\left(l(\xa)-l(\xx)+l(\xx)-l(\xh)\right) \mathbb{I} (\xh \leq \xx < \xa)]+\EX[(l(\xa)-l(\xx)) \mathbb{I} (\xx < \xh \leq \xa)] \notag\\
	& \overset{(b)}{\le} \EX[(l(\xa)-l(\xx)) \mathbb{I} (\xa \geq \xx)] -\EX[(l(\xh)-l(\xx)) \mathbb{I}(\xh \leq \xx < \xa)] \notag\\
	& \overset{(c)}{=}\EX[l(\xa) \mathbb{I} (\xa \geq \xx)] -\EX[l(\xh) \mathbb{I}(\xh \leq \xx < \xa)]-l(\xx) \PR((\xa=\xx) \cap (\xh>\xx, \xa>\xx)) \notag\\
	& \overset{(d)}{=} \EX[l(\xa) \mathbb{I} (\xa \geq \xx)] -\EX[l(\xh) \mathbb{I}(\xh \leq \xx < \xa)], \label{equ:MSE-combine-le-zero}
	\end{align}
	where $(a)$ holds by separating $l(\xa)-l(\xh)$ into $l(\xa)-l(\xx)+l(\xx)-l(\xh)$, $(b)$ follows from $\left(l(\xa)-l(\xx)\right) \mathbb{I} (\xh \leq \xx < \xa)+(l(\xa)-l(\xx)) \mathbb{I} (\xx < \xh \leq \xa) \le (l(\xa)-l(\xx)) \mathbb{I} (\xa \geq \xx)$, $(c)$ follows from $\mathbb{I} (\xa \geq \xx)-\mathbb{I}(\xh \leq \xx < \xa)=\mathbb{I} ((\xa=\xx) \cap (\xh>\xx, \xa>\xx))$, $(d)$ follows from $l(\xx) \ge 0$ due to Assumption \ref{ass:general-loss} (\romannumeral1).
	Thus, if $\eqref{equ:MSE-combine-le-zero} \leq 0$, we have $\eqref{equ:loss-two-intergral} < 0$ and $\EX[l(\xin)] > \EX[l(\xa)]$. So a necessary condition for $\EX[l(\xin)] \le \EX[l(\xa)]$ is $\eqref{equ:MSE-combine-le-zero} \ge 0$. Thus, we complete the proof of (\romannumeral3). \Halmos

\textbf{Proof of Example~\ref{prop:loss-counter-example}.}
	By the definition of $l(\cdot),\xa,\xh$, we have
	\begin{align}
	&\EX[l(\xa) \mathbb{I}(\xa \ge \xx)]=\sigma^2/2, \notag\\
	&\EX[l(\xh)\mathbb{I}(\xh \le \xx)]=\int_{-\infty}^{\xx-1} \frac{3 \epsilon}{(\sxh-\xx)^2} \dif \sxh= \int_{-\infty}^{-1} \frac{3 \epsilon}{\sxh^2} \dif \sxh=3 \epsilon. \notag 
	\end{align}
	So if $\epsilon < \sigma^2/(6a)$, then 
	\begin{equation*}
	\EX[l(\xa) \mathbb{I}(\xa \ge \xx)]=\sigma^2/2 \ge 3a \epsilon = a \EX[l(\xh)\mathbb{I}(\xh \le \xx)].
	\end{equation*}
	Next, we prove $\EX[l(\xin)] > \EX[l(\xa)]$. According to the distribution of $\xa, \xh$ \eqref{equ:loss-counter-example}, we have
	\begin{align*}
	&\EX[l(\xin)]-\EX[l(\xa)]\\
	&=\PR(\xh=\infty) \EX[(\xa-\xx)^2]+\int_{-\infty}^{\xx-1} \int_{-\infty}^{\infty} \left(\min\{\sxh,\sxa\}-\xx\right)^2 f(\sxa)\dif \sxa f(\sxh) \dif \sxh-\EX[(\xa-\xx)^2]\\
	&\overset{(a)}{=}\int_{-\infty}^{\xx-1}\int_{-\infty}^{\infty} \left(\min\{\sxh,\sxa\}-\xx\right)^2 f(\sxa)\dif \sxa f(\sxh) \dif \sxh-\epsilon \sigma^2\\
	&=\int_{-\infty}^{\xx-1}\int_{-\infty}^{\sxh} \left(\sxa-\xx\right)^2 f(\sxa)\dif \sxa f(\sxh) \dif \sxh+\int_{-\infty}^{\xx-1}\int_{\sxh}^{\infty} \left(\sxh-\xx\right)^2 f(\sxa)\dif \sxa f(\sxh) \dif \sxh-\epsilon \sigma^2\\
	& \overset{(b)}{\geq} \int_{-\infty}^{\xx-1}\int_{\sxh}^{\infty} \left(\sxh-\xx\right)^2 f(\sxa)\dif \sxa f(\sxh) \dif \sxh-\epsilon \sigma^2\\
	&=\int_{-\infty}^{\xx-1} \left(\sxh-\xx\right)^2 \int_{\sxh}^{\infty}  f(\sxa)\dif \sxa f(\sxh) \dif \sxh-\epsilon \sigma^2\\
	& \overset{(c)}{\geq} \frac{1}{2} \int_{-\infty}^{\xx-1}\left(\sxh-\xx\right)^2 f(\sxh) \dif \sxh-\epsilon \sigma^2\\
	&\overset{(d)}{=}\frac{1}{2} \int_{-\infty}^{\xx-1} \frac{3 \epsilon}{(\sxh-\xx)^2} \dif \sxh-\epsilon \sigma^2\\
	& = \frac{3 \epsilon}{2}-\epsilon \sigma^2\\
	& \overset{(e)}{>} 0,
	\end{align*}
	where $(a)$ follows from $\PR(\xh=\infty)=1-\epsilon$, $(b)$ follows from $\int_{-\infty}^{\xx-1}\int_{-\infty}^{\sxh} \left(\sxa-\xx\right)^2 f(\sxa)\dif \sxa f(\sxh) \dif \sxh \ge 0$, $(c)$ follows from  $\int_{\sxh}^{\infty} f(\sxa) \dif \sxa \ge \int_{0}^{\infty} f(\sxa) \dif \sxa = 1/2$ due to $\sxh \le \xx-1<0$, $(d)$ follows from the distribution of $\xh$ \eqref{equ:loss-counter-example}, $(e)$ follows from $\sigma^2 < 3/2$ in the condition of Example \ref{prop:loss-counter-example}. 
	\Halmos

\textbf{Proof of Corollary \ref{cor:loss-reduction}.}
	We define $D(\sxh) \coloneqq \EX[l(\xa)]-\EX[l(\xin)]$ as a function of $\sxh$. By \eqref{equ:MSE-minus}, we have
	\begin{equation}
	D(\sxh)-D(\sxh+\Delta)=\int_{\sxh}^{\sxh+\Delta} (l(\sxa)-l(\sxh)) f(\sxa) \dif \sxa,
	\end{equation}
	for any $\Delta>0$.
	
	When $\sxh \ge \xx$, we have $l(\sxa) \ge l(\sxh)$ for $\sxa \ge \sxh$ according to  Assumption \ref{ass:general-loss} (\romannumeral2). Thus, $D(\sxh) \ge D(\sxh+\Delta)$.
	
	When $\sxh < \xx$, there exists a small enough $\Delta$ such that $\sxh+\Delta < \xx$. And for $\sxh \le \sxa \le \sxh+\Delta<\xx$, we have $l(\sxa) \le l(\sxh)$ according to  Assumption \ref{ass:general-loss} (\romannumeral2). Thus, $D(\sxh) \le D(\sxh+\Delta)$. 
	
	In summary, $D(\sxh)$ increases as $\sxh$ when $\sxh < \xx$ and decreases when $\sxh \ge \xx$. \Halmos

\textbf{Proof of Proposition~\ref{prop:loss-two-side}.}
	We write down the difference of the expected losses:
	\begin{align}
	&\EX[l(\xa)]-\EX[l(\xin)] \notag\\
	&=\int_{-\infty}^{\infty} \int_{-\infty}^{\infty} \int_{-\infty}^{\infty} l(\sxa) f(\sxa,\sxhl,\sxhu) \dif \sxa \dif \sxhl \dif \sxhu-\int_{-\infty}^{\infty} \int_{-\infty}^{\infty} \int_{-\infty}^{\infty} l(\sxin) f(\sxa,\sxhl,\sxhu) \dif \sxa \dif \sxhl \dif \sxhu \notag\\
	&=\int_{-\infty}^{\infty} \int_{-\infty}^{\infty} \int_{-\infty}^{\sxhl} (l(\sxa)-l(\sxhl)) f(\sxa,\sxhl,\sxhu) \dif \sxa \dif \sxhl \dif \sxhu \notag\\
	&\quad + \int_{-\infty}^{\infty} \int_{-\infty}^{\infty} \int_{\sxhu}^{\infty} (l(\sxa)-l(\sxhu)) f(\sxa,\sxhl,\sxhu) \dif \sxa \dif \sxhl \dif \sxhu\notag\\
	&=\EX[(l(\xa)-l(\xhl)) \mathbb{I}(\xa \le \xhl)]+\EX[(l(\xa)-l(\xhu)) \mathbb{I}(\xa \ge \xhu)]
	, \label{equ:loss-two-side-dif}
	\end{align}
	where the second equality holds by the definition $\xin=\min\{\max\{\xa,\xhl\},\xhu\}$ and $\xhu \ge \xhl$.
	This proves part (\romannumeral1).
	
	Next, we prove (\romannumeral2).
	Separating the first term in \eqref{equ:loss-two-side-dif} by $\xhl < \xx$ and $\xhl \ge \xx$,
	we have
	\begin{align}
	&\EX[(l(\xa)-l(\xhl)) \mathbb{I}(\xa \le \xhl)] \notag\\
	&=\EX[(l(\xa)-l(\xhl)) \mathbb{I}(\xa \le \xhl < \xx)]+\EX[(l(\xa)-l(\xhl)) \mathbb{I}(\xa \le \xhl, \xhl \ge \xx)] \notag\\
	& \overset{(a)}{\ge} \EX[(l(\xa)-l(\xhl)) \mathbb{I}(\xa \le \xhl, \xhl \ge \xx)] \notag\\
	&\overset{(b)}{=}\EX\left[(l(\xa)-l(\xhl)) \mathbb{I}(\xa \le \xx, \xhl \ge \xx)\right]+\EX\left[(l(\xa)-l(\xhl)) \mathbb{I}(\xa > \xx, \xhl \ge \xa)\right] \notag\\
	&\overset{(c)}{=}\EX\left[(l(\xa)-l(\xhl)) \mathbb{I}(\xa \le \xx, \xhl \ge \xx)\right]+\EX\left[(l(\xa)-l(\xx)+l(\xx)-l(\xhl)) \mathbb{I}(\xa > \xx, \xhl \ge \xa)\right] \notag\\
	&\overset{(d)}{\ge}\EX\left[(l(\xa)-l(\xhl)) \mathbb{I}(\xa \le \xx, \xhl \ge \xx)\right]+\EX\left[(l(\xx)-l(\xhl)) \mathbb{I}(\xa > \xx, \xhl \ge \xa)\right] \notag\\
	& \overset{(e)}{\ge} \EX\left[(l(\xa)-l(\xx)+l(\xx)-l(\xhl)) \mathbb{I}(\xa \le \xx, \xhl \ge \xx)\right]+\EX\left[(l(\xx)-l(\xhl)) \mathbb{I}(\xa > \xx, \xhl \ge \xx)\right] \notag \\
	&\overset{(f)}{=}\EX\left[(l(\xa)-l(\xx)) \mathbb{I}(\xa \le \xx, \xhl \ge \xx)\right]+\EX\left[(l(\xx)-l(\xhl)) \mathbb{I}(\xhl \ge \xx)\right] \notag \\
	&\overset{(g)}{=}\EX\left[l(\xa) \mathbb{I}(\xa \le \xx, \xhl \ge \xx)\right]-\EX\left[l(\xhl) \mathbb{I}(\xhl \ge \xx)\right]+\EX[l(\xx) \mathbb{I}(\xa > \xx, \xhl \ge \xx)] \notag \\
	& \overset{(h)}{\ge} \EX\left[l(\xa) \mathbb{I}(\xa \le \xx, \xhl \ge \xx)\right]-\EX\left[l(\xhl) \mathbb{I}(\xhl \ge \xx)\right]
	, \label{equ:h1}
	\end{align}
	where the $(a)$ follows from $(l(\xa)-l(\xhl)) \mathbb{I}(\xa \le \xhl < \xx)$ due to Assumption \ref{ass:general-loss} (\romannumeral2),
	$(b)$ follows from separating the expectation by $\xa \le \xx$ and $\xa > \xx$, $(c)$  follows from $l(\xa)-l(\xhl)=l(\xa)-l(\xx)+l(\xx)-l(\xhl)$, $(d)$ follows from $\EX\left[(l(\xa)-l(\xx)) \mathbb{I}(\xa > \xx, \xhl \ge \xa)\right] \ge 0$ due to Assumption \ref{ass:general-loss} (\romannumeral2), $(e)$ follows from $\mathbb{I}(\xa \ge \xx, \xhl \ge \xa) \le \mathbb{I}(\xa \ge \xx, \xhl \ge \xx)$, $(f)$ follows from $\mathbb{I}(\xa \le \xx, \xhl \ge \xx)+\mathbb{I}(\xa > \xx, \xhl \ge \xx)=\mathbb{I}(\xhl \ge \xx)$, $(g)$ follows from $\mathbb{I}(\xhl \ge \xx)-\mathbb{I}(\xa \le \xx, \xhl \ge \xx)=\mathbb{I}(\xa > \xx, \xhl \ge \xx)$, $(h)$ follows from $l(\xx) \ge 0$ due to Assumption \ref{ass:general-loss} (\romannumeral1).
	
	For the second term in \eqref{equ:loss-two-side-dif}, we separate it by $\xhu \le \xx$ and $\xhu > \xx$:
	\begin{align}
	&\EX[(l(\xa)-l(\xhu)) \mathbb{I}(\xa \ge \xhu)] \notag\\
	&=\EX[(l(\xa)-l(\xhu)) \mathbb{I}(\xa \ge \xhu > \xx)]+\EX[(l(\xa)-l(\xhu)) \mathbb{I}(\xa \ge \xhu, \xx \ge \xhu)] \notag\\
	& \overset{(a)}{\ge} \EX[(l(\xa)-l(\xhu)) \mathbb{I}(\xa \ge \xhu, \xx \ge \xhu)] \notag\\
	&\overset{(b)}{=}\EX\left[(l(\xa)-l(\xhu)) \mathbb{I}(\xa \le \xx, \xhu \le \xa)\right]+\EX\left[(l(\xa)-l(\xhu)) \mathbb{I}(\xa > \xx, \xhu \le \xx)\right] \notag \\
	&=\EX\left[(l(\xa)-l(\xx)+l(\xx)-l(\xhu)) \mathbb{I}(\xa \le \xx, \xhu \le \xa)\right]+\EX\left[(l(\xa)-l(\xhu)) \mathbb{I}(\xa > \xx, \xhu \le \xx)\right] \notag \\
	& \overset{(c)}{\ge} \EX\left[(l(\xx)-l(\xhu)) \mathbb{I}(\xa \le \xx, \xhu \le \xa)\right]+\EX\left[(l(\xa)-l(\xx)+l(\xx)-l(\xhu)) \mathbb{I}(\xa > \xx, \xhu \le \xx)\right] \notag \\
	&=\EX\left[(l(\xa)-l(\xx)) \mathbb{I}(\xa \ge \xx, \xhu \le \xx)\right]-\EX\left[(l(\xhu)-l(\xx)) \mathbb{I}(\xhu \le \xx)\right] \notag \\
	&=\EX\left[l(\xa) \mathbb{I}(\xa \ge \xx, \xhu \le \xx)\right]-\EX\left[l(\xhu) \mathbb{I}(\xhu \le \xx)\right]+l(\xx) \PR(\xa < \xx, \xh \le \xx) \notag \\
	& \overset{(d)}{\ge} \EX\left[l(\xa) \mathbb{I}(\xa \ge \xx, \xhu \le \xx)\right]-\EX\left[l(\xhu) \mathbb{I}(\xhu \le \xx)\right]
	, \label{equ:h2}
	\end{align}
	where $(a)$ follows from $(l(\xa)-l(\xhu)) \mathbb{I}(\xa \ge \xhu > \xx) \ge 0$, $(b)$ follows from separating the expectation by $\xa \le \xx$ and $\xa > \xx$, $(b)$ follows from Assumption \ref{ass:general-loss} (\romannumeral2), $(c)$ follows from $(l(\xa)-l(\xx)+l(\xx)-l(\xhu)) \mathbb{I}(\xa \le \xx, \xhu \le \xa)\ge 0 $ due to Assumption \ref{ass:general-loss} (\romannumeral2), $(d)$ follows from $l(\xx) \ge 0$ due to Assumption \ref{ass:general-loss} (\romannumeral1).

	Thus, if $\eqref{equ:h1}+ \eqref{equ:h2}> 0$, we have \eqref{equ:loss-two-side-dif}$> 0$ and $\EX[l(\xa)] \ge \EX[l(\xin)]$. Thus, we complete the proof for (\romannumeral2).
	
	To prove (\romannumeral3), similar to \eqref{equ:h1}, we separate the first term in \eqref{equ:loss-two-side-dif} by $\xx \le \xa \le \xhl$, $\xa \le \xx \le \xhl$, and $\xa \le \xhl \le \xx$:
	\begin{align}
	&\EX[(l(\xa)-l(\xhl)) \mathbb{I}(\xa \le \xhl)] \notag\\
	&=\EX[(l(\xa)-l(\xhl)) \mathbb{I}(\xa \le \xhl \le \xx)]+\EX\left[(l(\xa)-l(\xhl)) \mathbb{I}(\xa \le \xx, \xhl \ge \xx)\right] \notag\\
	& \quad+\EX\left[(l(\xa)-l(\xhl)) \mathbb{I}(\xa \ge \xx, \xhl \ge \xa)\right] \notag\\
	&\overset{(a)}{\le} \EX[(l(\xa)-l(\xhl)) \mathbb{I}(\xa \le \xhl \le \xx)]+\EX\left[(l(\xa)-l(\xhl)) \mathbb{I}(\xa \le \xx, \xhl \ge \xx)\right] \notag\\
	&\overset{(b)}{=} \EX[(l(\xa)-l(\xx)+l(\xx)-l(\xhl)) \mathbb{I}(\xa \le \xx, \xhl \ge \xa)] \notag\\
	&\overset{(c)}{\le}
	\EX[(l(\xa)-l(\xx)) \mathbb{I}(\xa \le \xx)]+\EX[(l(\xx)-l(\xhl)) \mathbb{I}(\xa \le \xx \le \xhl)] \notag \\
	&=
	\EX[l(\xa)\mathbb{I}(\xa \le \xx)]-\EX[l(\xhl) \mathbb{I}(\xa \le \xx \le \xhl)]-l(\xx) \PR(\xa \le \xx, \xh > 
	\xx) \notag \\
	&\overset{(d)}{\le}
	\EX[l(\xa)\mathbb{I}(\xa \le \xx)]-\EX[l(\xhl) \mathbb{I}(\xa \le \xx \le \xhl)], \label{equ:loss-two-h1-up}
	\end{align}
	where $(a)$ follows from $(l(\xa)-l(\xhl)) \mathbb{I}(\xa \ge \xx, \xhl \ge \xa) \le 0$ due to Assumption \ref{ass:general-loss} (\romannumeral2), $(b)$ follows from $\mathbb{I}(\xa \le \xhl \le \xx)+\mathbb{I}(\xa \le \xx, \xhl \ge \xx)=\mathbb{I}(\xa \le \xx, \xhl \ge \xa)$, $(c)$ follows from $l(\xa)-l(\xx) \ge 0$ and $l(\xh)-l(\xx) \ge 0$ due to Assumption \ref{ass:general-loss} (\romannumeral2) and $\mathbb{I}(\xa \le \xx, \xhl \ge \xa) \ge \mathbb{I}(\xa \le \xx \le \xhl)$, $(d)$ follows from $l(\xx) \ge 0$ due to Assumption \ref{ass:general-loss} (\romannumeral1).
	
	Similar to \eqref{equ:h2}, we separate the second term in \eqref{equ:loss-two-side-dif} by $\xx \le \xhu \le \xa$, $\xhu \le \xx \le \xa$, and $\xhu \le \xa \le \xx$:
	\begin{align}
	&\EX[(l(\xa)-l(\xhu)) \mathbb{I}(\xa \ge \xhu)] \notag\\
	&=\EX[(l(\xa)-l(\xhu)) \mathbb{I}(\xa \ge \xhu \ge \xx)]+\EX\left[(l(\xa)-l(\xhu)) \mathbb{I}(\xa \le \xx, \xhu \le \xa)\right]\\
	&\quad+\EX\left[(l(\xa)-l(\xhu)) \mathbb{I}(\xa \ge \xx, \xhu \le \xx)\right] \notag\\
	&\overset{(a)}{\le} \EX[(l(\xa)-l(\xhu)) \mathbb{I}(\xa \ge \xhu \ge \xx)]+\EX\left[(l(\xa)-l(\xhu)) \mathbb{I}(\xa \ge \xx, \xhu \le \xx)\right] \notag\\
	&\overset{(b)}{=} \EX[(l(\xa)-l(\xx)+l(\xx)-l(\xhu)) \mathbb{I}(\xa \ge \xx, \xhu \le \xa)] \notag\\
	&\overset{(c)}{\le} \EX[(l(\xa)-l(\xx))\mathbb{I}(\xa \ge \xx)]+ \EX[(l(\xx)-l(\xhu))\mathbb{I}(\xa \ge \xx, \xhu \le \xx) ] \notag\\
	&=\EX[l(\xa)\mathbb{I}(\xa \ge \xx)]- \EX[l(\xhu)\mathbb{I}(\xa \ge \xx, \xhu \le \xx)]-l(\xx) \PR(\xa \ge \xx, \xhu > \xx) \notag\\ 
	&\overset{(d)}{\le} \EX[l(\xa)\mathbb{I}(\xa \ge \xx)]- \EX[l(\xhu)\mathbb{I}(\xa \ge \xx, \xhu \le \xx)], \label{equ:loss-two-h2-up}
	\end{align}
	where $(a)$ follows from $(l(\xa)-l(\xhu)) \mathbb{I}(\xa \le \xx, \xhu \le \xa) \le 0$ due to Assumption \ref{ass:general-loss} (\romannumeral2), $(b)$ follows from $\mathbb{I}(\xa \ge \xhu \ge \xx)+\mathbb{I}(\xa \ge \xx, \xhu \le \xx)=\mathbb{I}(\xa \ge \xx, \xhu \le \xa)$, $(c)$ follows from $l(\xa)-l(\xx) \ge 0$, $l(\xx)- l(\xh) \le 0$ and $\mathbb{I}(\xa \ge \xx, \xhu \le \xa) \ge \mathbb{I}(\xa \ge \xx, \xhu \le \xx)$, $(d)$ follows from $l(\xx) \ge 0$ due to Assumption \ref{ass:general-loss} (\romannumeral1). 
	
	
	Thus, if $\eqref{equ:loss-two-h1-up}+ \eqref{equ:loss-two-h2-up}< 0$, then \eqref{equ:loss-two-side-dif}$< 0$ and $\EX[l(\xa)] < \EX[l(\xin)]$. So a necessary condition for $\EX[l(\xin)] \le \EX[l(\xa)]$ is $\eqref{equ:loss-two-side-dif} \ge 0$. We complete the proof for (\romannumeral3). \Halmos

\textbf{Proof of Proposition \ref{prop:loss-two-side-cov}.}
	The proof basically follows the same argument as in Proposition~\ref{prop:loss-two-side}. We take the first part (\romannumeral1) as an example.
	We write down the difference of the expected losses:
	\begin{align}
	&\EX[l(\xa,W)]-\EX[l(\xin,W)] \notag\\
	&=\int_{-\infty}^{\infty}\int_{-\infty}^{\infty} \int_{-\infty}^{\infty}\int_{-\infty}^{\infty} l(\sxa,w) f(\sxa,\sxhl,\sxhu,w) \dif \sxa \dif \sxhl \dif \sxhu \dif w \notag\\
	&\quad -\int_{-\infty}^{\infty}\int_{-\infty}^{\infty} \int_{-\infty}^{\infty} \int_{-\infty}^{\infty} l(\sxin,w) f(\sxa,\sxhl,\sxhu,w)\dif \sxa \dif \sxhl \dif \sxhu \dif w \notag\\
	&=\int_{-\infty}^{\infty} \int_{-\infty}^{\infty}\int_{-\infty}^{\infty} \int_{-\infty}^{\sxhl} (l(\sxa,w)-l(\sxhl,w)) f(\sxa,\sxhl,\sxhu,w)\dif \sxa \dif \sxhl \dif \sxhu \dif w \notag\\
	&\quad+\int_{-\infty}^{\infty} \int_{-\infty}^{\infty} \int_{-\infty}^{\infty}\int_{\sxhu}^{\infty} (l(\sxa,w)-l(\sxhu,w)) f(\sxa,\sxhl,\sxhu,w)\dif \sxa \dif \sxhl \dif \sxhu \dif w \notag\\
	&=\EX[(l(\xa,W)-l(\xhl,W)) \mathbb{I}(\xa \le \xhl)]+\EX[(l(\xa,W)-l(\xhu,W)) \mathbb{I}(\xa \ge \xhu)]. \notag 
	\end{align} 
	Thus, we complete the proof for part (\romannumeral1). \Halmos

 
\section{Proofs in Section \ref{sec:compe-pricing}}
\textbf{Proof of Lemma \ref{lem:compe-AI-price}.}
	We reformulate the linear demand function \eqref{compe-true-demand} in the matrix form:
	\begin{equation*}
	D=A \theta+\gamma \bm{p'}+\bm{\epsilon},
	\end{equation*}
	where 
	\begin{equation*}
	D=
	\begin{pmatrix}
	d_1\\
	\vdots\\
	d_n
	\end{pmatrix}, \ 
	A=
	\begin{pmatrix}
	1 & -p_1\\
	\vdots & \vdots\\
	1 & -p_n
	\end{pmatrix}, \ 
	\theta=\begin{pmatrix}
	\alpha\\
	\beta\\
	\end{pmatrix}, \
	\bm{p'}=\begin{pmatrix}
	p_1'\\
	\vdots\\
	p_n'
	\end{pmatrix}, \ 
	\bm{\epsilon}=\begin{pmatrix}
	\epsilon_1\\
	\vdots\\
	\epsilon_n
	\end{pmatrix}.
	\end{equation*}
	According to the algorithm assumed demand model \eqref{equ:compe-demand-AI}, the OLS estimator is 
	\begin{equation}\label{equ:compe-OLS-pricing}
	\hat{\theta}=
	\begin{pmatrix}
	\hat{\alpha} \\ \hat{\beta}
	\end{pmatrix}
	=(A^\top A)^{-1} A^\top D = \theta + (A^\top A)^{-1} A^\top \bm{p'} \gamma + (A^\top A)^{-1} A^\top \epsilon.
	\end{equation}
	According to Assumption \ref{asp:comp-pricing}, we have 
	\begin{align}
	&\frac{1}{n} \sum_{i=1}^n p_i \overset{p}{\longrightarrow} \mu, \ \frac{1}{n} \sum_{i=1}^n p_i' \overset{p}{\longrightarrow} \mu, \label{equ:compe-mean-converge}\\
	&\frac{1}{n} \sum_{i=1}^n p_i^2 \overset{p}{\longrightarrow} \mu^2+\sigma^2, \ \frac{1}{n} \sum_{i=1}^n p_i'^2 \overset{p}{\longrightarrow} \mu^2+\sigma^2, \label{equ:compe-var-converge}\\
	&\frac{1}{n} \sum_{i=1}^n p_i p_i'-\mu^2\overset{p}{\longrightarrow} \rho \sigma^2,\label{equ:compe-cov-converge}
	\end{align}
	where $\overset{p}{\longrightarrow}$ denotes convergence in probability, the convergence follows by the weak law of large numbers. 
	By~\eqref{equ:compe-mean-converge}, \eqref{equ:compe-var-converge}, and  \eqref{equ:compe-cov-converge}, we have
	\begin{equation*}
	\frac{1}{n} A^\top A = 
	\begin{pmatrix}
	1 & -\frac{1}{n} \sum_{i=1}^n p_i \\ -\frac{1}{n} \sum_{i=1}^n p_i  & \frac{1}{n} \sum_{i=1}^n p_i^2
	\end{pmatrix} \overset{p}{\longrightarrow}
	\begin{pmatrix}
	1 & -\mu\\
	-\mu & \mu^2+\sigma^2
	\end{pmatrix},
	\end{equation*}
	and
	\begin{equation*}
	\frac{1}{n} A^\top \bm{p'} =
	\begin{pmatrix}
	\frac{1}{n} \sum_{i=1}^n p_i' \\ 
	-\frac{1}{n} \sum_{i=1}^n p_i p_i'
	\end{pmatrix}
	\overset{p}{\longrightarrow}
	\begin{pmatrix}
	\mu\\
	-(\rho \sigma^2+\mu^2)
	\end{pmatrix}.
	\end{equation*}
	And by Slutsky's theorem, we have
	\begin{equation*}
	(\frac{1}{n} A^\top A)^{-1} \frac{1}{n} A^\top \bm{p'} \gamma \overset{p}{\longrightarrow} \frac{\gamma}{\sigma^2}
	\begin{pmatrix}
	\sigma^2+\mu^2 & \mu\\
	\mu & 1\\
	\end{pmatrix}
	\begin{pmatrix}
	\mu \\ -\rho \sigma^2-\mu^2
	\end{pmatrix}
	\!=\!\frac{\gamma}{\sigma^2}
	\begin{pmatrix}
	\mu \sigma^2 + \mu^3 -\mu \rho \sigma^2 -\mu^3\\
	- \rho \sigma^2
	\end{pmatrix}
	\!=\!\begin{pmatrix}
	\gamma \mu (1-\rho)\\
	-\gamma \rho.
	\end{pmatrix}
	\end{equation*}
	Since $\epsilon$ is the random noise, we have $ (A^\top A)^{-1} A^\top \epsilon \overset{p}{\longrightarrow} 0$.
	
	So the OLS estimator in \eqref{equ:compe-OLS-pricing} is 
	\begin{equation*}
	\hat{\theta}=
	\begin{pmatrix}
	\hat{\alpha} \\ \hat{\beta}
	\end{pmatrix} 
	=\begin{pmatrix}
	\alpha\\
	\beta
	\end{pmatrix}
	+\begin{pmatrix}
	\gamma \mu (1-\rho)\\
	-\gamma \rho.
	\end{pmatrix}
	\overset{p}{\longrightarrow}
	\begin{pmatrix}
	\alpha+ \gamma \mu (1-\rho)\\
	\beta-\gamma \rho.
	\end{pmatrix}
	\end{equation*}
	Then the optimal price for AI is
	\begin{equation*}
	\pa=\frac{\hat{\alpha}}{2 \hat{\beta}} \overset{p}{\longrightarrow} \frac{\alpha+ \gamma \mu (1-\rho)}{2(\beta-\gamma \rho)},
	\end{equation*}
	which completes the proof for Lemma \ref{thm:compe-pricing}. \Halmos

\textbf{Proof of Theorem \ref{thm:compe-pricing}.}
	We define the revenue function $r(p)\coloneqq pd(p,p')$ and provide a condition for $\prin$ such that $r(\prin) \ge r(\pa)$. Note that $\prin=\min\{\pa,\ph\}$. If $\pa \le \ph$, then $r(\prin) = r(\pa)$. So the condition for $\prin$ boils down to the condition for $\ph$ such that $r(\ph) \ge r(\pa)$ when $\ph < \pa$. According to \eqref{equ:compe-demand-AI}, we have 
	\begin{align}
	&r(\ph)-r(\pa)\notag\\
	&=\ph(\alpha+(\gamma-\beta)\ph)-\pa(\alpha-\beta \pa+\gamma \ph) \notag\\
	&=(\gamma-\beta) \ph^2 + \left(\alpha-\frac{\gamma}{2} \frac{\alpha+\gamma \mu (1-\rho)}{\beta-\rho \gamma}\right) \ph+\frac{\beta}{4} \frac{(\alpha+\gamma \mu (1-\rho))^2}{(\beta-\rho \gamma)^2}- \frac{\alpha}{2} \frac{\alpha+\gamma \mu (1-\rho)}{\beta- \rho \gamma}. \label{equ:compe-price-revenue-dif}
	\end{align}
	So $r(\ph) \ge r(\pa)$ is equivalent to 
	\eqref{equ:compe-price-revenue-dif} $\ge 0$. And multiplying \eqref{equ:compe-price-revenue-dif} by $-4(\beta-\rho \gamma)^2$, we have 
	\begin{align}
	&4(\beta-\rho \gamma)^2(\beta-\gamma) \ph^2 -2 (\beta-\rho \gamma) \left(2 \alpha \beta-2 \alpha \rho \gamma - \alpha \gamma -\gamma^2 \mu(1-\rho)\right) \ph \notag\\
	&\quad-\left(\alpha+\gamma \mu (1-\rho)\right) \left(\beta \gamma \mu (1-\rho) +2 \alpha \rho \gamma - \alpha \beta\right) \le 0. \label{equ:compe-price-function}
	\end{align}
	We write down the discriminant of the quadratic function \eqref{equ:compe-price-function}:
	\begin{align}\label{equ:compe-price-discriminant}
	\Delta&=4(\beta-\rho \gamma)^2 (2 \alpha \beta-2 \alpha \gamma \rho - \alpha \gamma - \gamma^2 \mu+ \gamma^2 \mu \rho)^2  \notag\\
	&\quad+ 16 (\beta-\rho \gamma)^2 (\beta-\gamma)(\alpha+\gamma \mu(1-\rho)) \left(\beta \gamma \mu (1-\rho)+2 \alpha \rho \gamma-\alpha \beta\right),
	\end{align}
	In the first term of \eqref{equ:compe-price-discriminant}, we have 
	\begin{align}
	&(2 \alpha \beta-2 \alpha \gamma \rho - \alpha \gamma - \gamma^2 \mu+ \gamma^2 \mu \rho)^2 \notag\\
	&=4 \alpha^2 \beta^2 +4 \alpha^2 \gamma^2 \rho^2 +\alpha^2 \gamma^2+\gamma^4 \mu^2+\gamma^4 \mu^2 \rho^2-8 \alpha^2 \beta \gamma \rho-4 \alpha^2 \beta \gamma -4 \alpha \beta \gamma^2 \mu+4 \alpha \beta \gamma^2 \mu \rho+4 \alpha^2 \gamma^2 \rho \notag\\
	&\quad+4 \alpha \gamma^3 \mu \rho -4 \alpha \gamma^3 \mu \rho^2+2\alpha \gamma^3 \mu -2 \alpha \gamma^3 \mu \rho -2 \gamma^4 \mu^2 \rho. \label{equ:compe-price-discriminant-one}
	\end{align}
	In the second term of \eqref{equ:compe-price-discriminant}, we have
	\begin{align}
	&(\beta-\gamma)(\alpha+\gamma \mu(1-\rho)) \left(\beta \gamma \mu (1-\rho)+2 \alpha \rho \gamma-\alpha \beta\right) \notag\\
	&=\!-\!\alpha^2 \beta^2\!+\!2 \alpha^2 \beta \gamma \rho \!+\!2 \alpha \beta \gamma^2 \mu(1-\rho) \rho\!+\!\beta^2 \gamma^2 \mu^2 (1-\rho)^2 \!+\! \alpha^2 \beta \gamma \!-\!2 \alpha^2 \gamma^2 \rho \!-\!2 \alpha \gamma^3 \mu (1-\rho) \rho\!-\! \beta \gamma^3 \mu^2 (1-\rho)^2.\label{equ:compe-price-discriminant-two}
	\end{align}
	Plugging \eqref{equ:compe-price-discriminant-one} and \eqref{equ:compe-price-discriminant-two} into \eqref{equ:compe-price-discriminant}, we have 
	\begin{equation*}
	\frac{\Delta}{4(\beta-\rho \gamma)^2}=\gamma^2 \left(\alpha(1-2\rho)+(\gamma-2\beta)(1-\rho)\mu\right)^2,
	\end{equation*}
	and the roots of \eqref{equ:compe-price-function}
	\begin{equation}\label{equ:compe-price-root}
	p=\dfrac{2 \alpha \beta- 2 \alpha \rho \gamma -\alpha \gamma - \gamma^2 \mu(1-\rho) \pm \Big| \gamma \left(\alpha(1-2 \rho)+(\gamma-2 \beta)(1-\rho) \mu\right)\Big|}{4(\beta-\rho \gamma)(\beta-\gamma)}.
	\end{equation}
	Let the function $h(\rho) \coloneqq \alpha(1-2 \rho)+(\gamma-2 \beta)(1-\rho) \mu$. Since we assume $\mu \ge p_{NE}=\frac{\alpha}{2 \beta-\gamma}$, we have  
	\begin{equation*}
	h(0)=\alpha+(\gamma-2 \beta) \mu \le 0, \ h(1)=-\alpha <0.
	\end{equation*}
	Since $h(\rho)$ is a linear function of $\rho$, we have $h(\rho) \le 0$ for any $\rho \in [0,1]$.
	So the roots of \eqref{equ:compe-price-revenue-dif} are 
	\begin{equation*}
	p_L=\frac{\alpha \beta - 2\alpha \gamma \rho - \beta \gamma (1-\rho) \mu}{2(\beta-\rho \gamma)(\beta-\gamma)}, \ p_H=\frac{\alpha+ \gamma \mu (1-\rho)}{2(\beta-\gamma \rho)}.
	\end{equation*}
	Note that $p_L < p_H =\pa$.
	Then the sufficient condition for $\eqref{equ:compe-price-revenue-dif} \ge 0$ is that $ p_L \le \ph \le \pa$. 
	Especially, if $p_L<\ph <\pa$, $r(\prin)$ is strictly greater than $r(\pa)$.
	
	Next, we prove that $p_L \le p_{NE} \le p_H$. First, we consider
	\begin{align}
	p_{NE}-p_L
	&=\frac{\alpha}{2 \beta-\gamma}-\frac{\alpha \beta - 2\alpha \gamma \rho - \beta \gamma (1-\rho) \mu}{2(\beta-\rho \gamma)(\beta-\gamma)} \notag\\
	&=\dfrac{2\beta \mu-\alpha-\gamma \mu+(2 \alpha-2 \beta \mu +\gamma \mu)\rho}{2(2\beta-\gamma)(\beta-\rho \gamma)(\beta-\gamma)}. \label{equ:compe-price-NEL}
	\end{align}
	Since $\beta > \gamma$, $\rho \in [0,1]$, we have the denominator in \eqref{equ:compe-price-NEL} is greater than zero. 
	We claim that the numerator in \eqref{equ:compe-price-NEL} is greater than zero for all $\rho \in [0,1]$. To see this, let the linear function 
	\begin{equation*}
	g(\rho) \coloneqq 2\beta \mu-\alpha-\gamma \mu+(2 \alpha-2 \beta \mu +\gamma \mu)\rho.
	\end{equation*}
	We have $g(0)=2\beta \mu-\alpha-\gamma \mu \ge 0$ because of $\mu \ge p_{NE}=\frac{\alpha}{2 \beta-\gamma}$. Also, we have $g(1)=\alpha >0$. Thus, for any $\rho \in [0,1]$, we have $g(\rho) \ge 0$. So $\eqref{equ:compe-price-NEL} \ge 0$ and $p_{NE} \ge p_L$.
	Next, we have 
	\begin{equation}\label{equ:compe-price-NEH}
	p_H-p_{NE}=\dfrac{g(\rho)}{2(2\beta-\gamma)(\beta-\rho \gamma)}=\dfrac{2\beta \mu-\alpha-\gamma \mu+(2 \alpha-2 \beta \mu +\gamma \mu)\rho}{2(2\beta-\gamma)(\beta-\rho \gamma)} \ge 0.
	\end{equation} 
	Combining \eqref{equ:compe-price-NEL} and \eqref{equ:compe-price-NEH}, we have $p_L \le p_{NE} \le p_H$. \Halmos

\section{Proofs in Section \ref{sec:mis}}

\textbf{Proof of Proposition \ref{prop:mis-finite-human}.}
	We first prove (\romannumeral1), the finite-sample result for the algorithmic decision.
	The OLS estimator is
	\begin{equation}\label{equ:mis-OLS-finite}
	\hat{\theta}=(A^\top A)^{-1} A^\top(f(\bm{p})+\bm{\epsilon}),
	\end{equation} 
	where $A$ is the design matrix 
	\begin{equation*}
	A=\begin{pmatrix}
	A_0 \\ A_1 \\ \vdots \\A_n
	\end{pmatrix}_{(n+1)K \times 2},\ 
	A_j=\begin{pmatrix}
	1 & -p_j \\
	1 & -p_j \\
	\vdots\\
	1 & -p_j
	\end{pmatrix}_{K \times 2},\ 
	f(\bm{p})=\begin{pmatrix}
	f(p_0) \\ f(p_0) \\ \vdots \\ f(p_n)
	\end{pmatrix}_{(n+1)K \times 1},\ 
	\bm{\epsilon}=\begin{pmatrix}
	\epsilon_{01} \\ \epsilon_{01} \\ \vdots \\ \epsilon_{nK}
	\end{pmatrix}_{(n+1)K \times 1}.
	\end{equation*}
	Then, we have
	\begin{align}\label{equ:mis-design-A}
	\frac{1}{(n+1)K} A^\top A
	&=\frac{1}{(n+1)K} 
	\begin{pmatrix}
	A_0^\top & A_1^\top \cdots & A_n^\top
	\end{pmatrix}
	\begin{pmatrix}
	A_0 \\ A_1 \\ \vdots \\ A_n
	\end{pmatrix}
	=\frac{1}{(n+1)K} \sum_{i=0}^n A_i^\top A_i \notag \\
	&=\frac{1}{n+1}\sum_{i=0}^n 	\begin{pmatrix}
	1 \\ - p_i
	\end{pmatrix}
	\begin{pmatrix}
	1 & - p_i\\
	\end{pmatrix}
	=\frac{1}{n+1} \sum_{i=0}^n 
	\begin{pmatrix}
	1 & - p_i\\
	- p_i &  p_i^2
	\end{pmatrix}.
	\end{align}
	According to the price grid \eqref{equ:mis-discrete-prices}, we have
	\begin{align}
	& \frac{1}{n+1} \sum_{i=0}^n p_i=\frac{1}{n+1} \left(c(n+1)+\sum_{i=0}^n i \frac{\bar{p}-c}{n} \right)=\frac{1}{2} (\bar{p}+c), \label{equ:mis-pi-aver}\\
	& \frac{1}{n+1} \sum_{i=0}^n p_i^2=\frac{1}{n+1} \left(\sum_{i=0}^n c^2+2i c\frac{\bar{p}-c}{n}+i^2 \frac{(\bar{p}-c)^2}{n^2} \right)=c\bar{p}+\frac{2n+1}{6n} (\bar{p}-c)^2.\label{equ:mis-pi2-aver}
	\end{align}
	Plugging \eqref{equ:mis-pi-aver} and \eqref{equ:mis-pi2-aver} into \eqref{equ:mis-design-A}, we have
	\begin{align}
	&\frac{1}{(n+1)K} A^\top A=
	\begin{pmatrix}
	1 & -\frac{1}{2}(\bar{p}+c) \\
	-\frac{1}{2}(\bar{p}+c) & c\bar{p}+\frac{2n+1}{6n} (\bar{p}-c)^2
	\end{pmatrix}, \notag \\
	&\left(\frac{1}{(n+1)K} A^\top A\right)^{-1}=\frac{1}{\frac{n+2}{12n}(\bar{p}^2+c^2)-\frac{n+2}{6n} \bar{p}c}
	\begin{pmatrix}
	c\bar{p}+\frac{2n+1}{6n} (\bar{p}-c)^2 & \frac{1}{2}(\bar{p}+c) \\
	\frac{1}{2}(\bar{p}+c) & 1
	\end{pmatrix},\label{equ:mis-def-AA-inverse} \\
	& \frac{1}{(n+1)K}A^\top (f(\bm{p})+\bm{\epsilon})=
	\begin{pmatrix}
	\frac{1}{n+1} \sum_{i=0}^n f(p_i)+\frac{1}{(n+1)K} \sum_{i=0}^{n} \sum_{j=1}^K \epsilon_{ij} \\
	-\frac{1}{n+1} \sum_{i=0}^n p_i f(p_i)-\frac{1}{(n+1)K} \sum_{i=0}^n p_i \sum_{j=1}^K \epsilon_{ij}
	\end{pmatrix}
	\coloneqq
	\begin{pmatrix}
	c_{1n}\\
	-c_{2n}
	\end{pmatrix}
	.\label{equ:mis-def-c1-c2}
	\end{align}
	Plugging \eqref{equ:mis-def-AA-inverse} and \eqref{equ:mis-def-c1-c2} into \eqref{equ:mis-OLS-finite}, we have
	\begin{equation*}
	\hat{\theta}=\frac{1}{\frac{n+2}{12n}(\bar{p}^2+c^2)-\frac{n+2}{6n} \bar{p}c}
	\begin{pmatrix}
	c\bar{p}c_{1n}+\frac{2n+1}{6n} (\bar{p}-c)^2 c_{1n}-\frac{\bar{p}+c}{2} c_{2n}\\
	\frac{\bar{p}+c}{2} c_{1n} -c_{2n}
	\end{pmatrix},
	\end{equation*}
	where $c_{1n},c_{2n}$ are defined in $\eqref{equ:mis-def-c1-c2}$.
	The optimal price prescribed by AI is 
	\begin{equation}\label{equ:mis-LR-solution-finite}
	\pa=\frac{\hat{\alpha}}{2 \hat{\beta}}+\frac{c}{2}
	=\dfrac{c\bar{p}c_{1n}+\frac{2n+1}{6n} (\bar{p}-c)^2 c_{1n}-\frac{\bar{p}+c}{2} c_{2n}}{(\bar{p}+c) c_{1n}-2 c_{2n}}+\frac{c}{2}.
	\end{equation}
	Let 
	$c_1,c_2$ denote the estimator when $K \rightarrow \infty$:
	\begin{equation}
	\begin{pmatrix}
	c_1\\
	c_2
	\end{pmatrix}
	\coloneqq
	\begin{pmatrix}
	\frac{1}{n+1} \sum_{i=0}^n f(p_i)\\
	\frac{1}{n+1} \sum_{i=0}^n p_i f(p_i)
	\end{pmatrix}.
	\end{equation}
	
	In the first step, we will show that for small constants $\delta_1,\delta_2>0$,
	\begin{equation}\label{equ:mis-CI-c1-c2}
	|c_{1n}-c_1| \leq \delta_1, \ |c_{2n}-c_2| \le \frac{\bar{p}+c}{2} \delta_2,
	\end{equation}
	with a high probability. In the second step, since the AI price $\pa$ \eqref{equ:mis-LR-solution-finite} is continuous in $c_{1n},c_{2n}$, we have $|\pa-\pas| \le \delta$ with a high probability for a small constant $\delta$.
	
	\textbf{Step one:} Note that $c_{1n}-c_1$ is the average of $(n+1)K$ i.i.d. $\sigma$-sub-Gaussian variables:
	\begin{equation*}
	c_{1n}-c_1=\frac{1}{(n+1)K} \sum_{i=0}^{n} \sum_{j=1}^K \epsilon_{ij}.
	\end{equation*}
	By the concentration inequality (see Proposition 2.6.1 in \citealt{vershynin2018high}), we have 
	\begin{align}\label{equ:mis-c1n-con}
	\PR\left(|c_{1n}-c_1| \ge \delta_1\right) \le 2\exp\left(-\dfrac{\delta_1^2 (n+1)K}{2 \sigma^2}\right).
	\end{align}
	Similarly, we have 
	\begin{equation*}
	c_{2n}-c_2=\frac{1}{(n+1)K} \sum_{i=0}^{n} p_i \sum_{j=1}^K \epsilon_{ij}.
	\end{equation*}
	By the definition of $p_i$, we have
	\begin{equation*}
	\sum_{i=0}^n p_i^2 = (n+1) \left(c\bar{p}+\frac{2n+1}{6n} (\bar{p}-c)^2\right) \le \frac{1}{2} (n+1) (\bar{p}+c)^2,
	\end{equation*}
	where the inequality follows by $\frac{2n+1}{6n} \le 0.5$ due to $n \ge 1$.
	Thus, $c_{2n}-c_2$ is $\sqrt{\frac{(\bar{p}+c)^2}{2(n+1)K}} \sigma$-sub-Gaussian and 
	\begin{equation}\label{equ:mis-c2n-con}
	\PR\left(|c_{2n}-c_2| \ge \frac{\bar{p}+c}{2} \delta_2\right) \le 2\exp\left(-\frac{\delta_2^2(n+1)K}{4 \sigma^2} \right).
	\end{equation}
	
	\textbf{Step two:} Taking the partial derivative of $\pa$ with respect to $c_{1n}, c_{2n}$, we have
	\begin{align}
	&\frac{\partial \pa}{\partial c_{1n}}=\dfrac{-\left(\frac{1}{6}+\frac{1}{3n}\right) (\bar{p}-c)^2 c_{2n}}{\left((\bar{p}+c)c_{1n}-2c_{2n}\right)^2}\\
	&\frac{\partial \pa}{\partial c_{2n}}=\dfrac{\left(\frac{1}{6}+\frac{1}{3n}\right) (\bar{p}-c)^2 c_{1n}}{\left((\bar{p}+c)c_{1n}-2c_{2n}\right)^2}.
	\end{align}
	We set $\delta_1<c_1, \delta_2<\frac{2 c_2}{\bar{p}+c}$ in \eqref{equ:mis-CI-c1-c2} to make sure $c_{1n},c_{2n} > 0$. Thus, we have $\frac{\partial \pa}{\partial c_{1n}} < 0$ and $\frac{\partial \pa}{\partial c_{2n}} > 0$. 
	On the one hand, when $c_{1n},c_{2n}$ satisfy \eqref{equ:mis-CI-c1-c2}, the AI price $p_a$ attains the maximum when $c_{1n}=c_1-\delta_1$,
	$c_{2n}=c_2+\frac{\bar{p}+c}{2} \delta_2$. So we have an upper bound for $\pa$ by plugging $c_{1n},c_{2n}$ into \eqref{equ:mis-LR-solution-finite}:
	\begin{align}\label{equ:mis-LR-finite}
	\pa \le
	\dfrac{ \left(c\bar{p}+\frac{2n+1}{6n} (\bar{p}-c)^2\right) (c_1-\delta_1)-\frac{\bar{p}+c}{2} \left(c_2+\frac{\bar{p}+c}{2} \delta_2\right)}{(\bar{p}+c) (c_1-\delta_1)-2 \left(c_2+\frac{\bar{p}+c}{2} \delta_2\right)}+\frac{c}{2}.
	\end{align}
	If $K \rightarrow \infty$, we have $\delta_1 \rightarrow 0$, $\delta_2 \rightarrow 0$, and the AI price $\pa$ converging to
	\begin{equation}\label{equ:mis-LR-solution}
		\pas
		=\dfrac{c\bar{p}c_1+\frac{2n+1}{6n} (\bar{p}-c)^2 c_1-\frac{\bar{p}+c}{2} c_2}{(\bar{p}+c) c_1-2 c_2}+\frac{c}{2}.
	\end{equation}
	
	Let $A \coloneqq c\bar{p}+\frac{2n+1}{6n} (\bar{p}-c)^2$, $B \coloneqq \frac{\bar{p}+c}{2}$. By \eqref{equ:mis-LR-finite} and \eqref{equ:mis-LR-solution}, we have
	\begin{equation}
	\pa \le \dfrac{A c_1-B c_2 -A \delta_1 -B^2 \delta_2}{2 B c_1-2c_2-2B (\delta_1+\delta_2)}+\frac{c}{2}, \ \pas=\dfrac{A c_1-B c_2}{2 B c_1-2c_2}+\frac{c}{2}. 
	\end{equation}
	Thus,
	\begin{align}\label{equ:mis-gap-pa}
	\pa-\pas \le \dfrac{2(A-B^2)(c_2 \delta_1+Bc_1 \delta_2)}{\left(2 B c_1-2c_2-2B (\delta_1+\delta_2)\right)\left(2 B c_1-2c_2\right)},
	\end{align}
	where $A-B^2=\left(\frac{1}{12}+\frac{1}{6n}\right)(\bar{p}-c)^2$.
	We let 
	\begin{equation}\label{equ:mis-value-delta12}
	\delta_1=\frac{2(Bc_1-c_2)^2}{4 c_2 (A-B^2)} \delta=\dfrac{3\left((\bar{p}+c)c_1/2-c_2\right)^2}{c_2(1/2+1/n)(\bar{p}-c)^2} \delta, \ \delta_2=\frac{2(Bc_1-c_2)^2}{4 B c_1 (A-B^2)}\delta=\dfrac{6\left((\bar{p}+c)c_1/2-c_2\right)^2}{(\bar{p}+c)c_1(1/2+1/n)(\bar{p}-c)^2} \delta.
	\end{equation}
	For a constant $\delta$ satisfying
	\begin{equation}\label{equ:mis-con-delta2}
	\delta \le \dfrac{c_1 c_2 (A-B^2)}{(Bc_1+c_2)(Bc_1-c_2)},
	\end{equation}
	we can check $\delta_1<c_1, \delta_2<\frac{2 c_2}{\bar{p}+c}$ and 
	\begin{equation}\label{equ:mis-gap-denomi}
	2B(\delta_1+\delta_2) \le B c_1 -c_2.
	\end{equation}
	By \eqref{equ:mis-con-delta2}, \eqref{equ:mis-gap-pa}, \eqref{equ:mis-value-delta12} and \eqref{equ:mis-gap-denomi}, we have
	\begin{equation}
	\pa-\pas \le \delta.
	\end{equation}
	On the other hand, when $c_{1n},c_{2n}$ satisfy \eqref{equ:mis-CI-c1-c2}, the AI price $p_a$ attains the minimum when $c_{1n}=c_1+\delta_1$,
	$c_{2n}=c_2-\frac{\bar{p}+c}{2} \delta_2$. So we have
	\begin{align}\label{equ:mis-gap-pa-low}
	\pa-\pas \ge \dfrac{2(B^2-A)(c_2 \delta_1+Bc_1 \delta_2)}{\left(2 B c_1-2c_2-2B (\delta_1+\delta_2)\right)\left(2 B c_1-2c_2\right)}.
	\end{align}
	By \eqref{equ:mis-con-delta2}, \eqref{equ:mis-gap-pa-low}, \eqref{equ:mis-value-delta12} and \eqref{equ:mis-gap-denomi}, we have
	\begin{equation}
	\pa-\pas \ge -\delta.
	\end{equation}
	In summary, by \eqref{equ:mis-c1n-con}, \eqref{equ:mis-c2n-con}, we have
	\begin{equation*}
	\PR(|\pa-\pas| \ge \delta) \le 2\exp\left(-\dfrac{\delta_1^2 (n+1)K}{2 \sigma^2}\right)+ 2\exp\left(-\frac{\delta_2^2(n+1)K}{4 \sigma^2}\right).
	\end{equation*}

	Next, we prove (\romannumeral2), the finite-sample result for the range given by the human analyst. According to the property of strong concavity, we have the revenue $r(p) \coloneqq (p-c) f(p)$ is unimodal and $r'(p_j)>0$ for $p_j < p^*$. Thus, we have
	\begin{equation}\label{equ:mis-increase-p}
	r(p_{j-1}) \le r(p_j) - r'(p_j) \frac{\bar{p}}{n} -\frac{\lambda}{2} \left(\frac{\bar{p}-c}{n}\right)^2 < r(p_j) -\frac{\lambda}{2} \left(\frac{\bar{p}-c}{n}\right)^2,
	\end{equation} 
	for $p_j < p^*$.
	Similarly, for $p_j \ge p^*$, due to $r'(p_j) <0$, we have 
	\begin{equation}\label{equ:mis-decrease-p}
	r(p_{j+1}) \le r(p_j) + r'(p_j) \frac{\bar{p}}{n} -\frac{\lambda}{2} \left(\frac{\bar{p}-c}{n}\right)^2 < r(p_j) -\frac{\lambda}{2} \left(\frac{\bar{p}-c}{n}\right)^2.
	\end{equation}
	Let $\Delta \coloneqq \frac{\lambda}{4} \left(\frac{\bar{p}-c}{n}\right)^2$. Due to \eqref{equ:mis-increase-p}, \eqref{equ:mis-decrease-p}, we have
	\begin{align}\label{equ:mis-unimodal-p}
	r(p_{j-1})+\Delta < r(p_j)-\Delta \ \text{for} \ p_j<p^*, \ r(p_{j})-\Delta > r(p_{j+1})+\Delta \ \text{for} \ p_j \ge p^*.
	\end{align}
	By the concentration inequality of sub-Gaussain variables, we have 
	\begin{align*}
	\PR(|\hat{r}(p_j)-r(p_j)| \ge \Delta)
	&=2\PR\left(\frac{1}{K} \sum_{k=1}^K \epsilon_{jk} \ge \frac{\Delta}{p_j}\right)
	\le 2\PR\left(\frac{1}{K} \sum_{k=1}^K \epsilon_{jk} \ge \frac{\Delta}{\bar{p}}\right)\\
	&\le 2 \exp\left(-\frac{K \Delta^2}{2 \sigma^2 \bar{p}^2 }\right)=2 \exp\left(-\frac{K \lambda^2 (\bar{p}-c)^4}{32 \sigma^2 \bar{p}^2 n^4}\right).
	\end{align*}
	Define the good event $G=\left\{|\hat{r}(p_j)-r(p_j)| < \Delta, \ \forall j \in [0,1,\ldots,n]\right\}$. We have
	\begin{equation*}
	\PR(G) \ge 1- 2(n+1) \exp\left(-\frac{K \lambda^2 (\bar{p}-c)^4}{32 \sigma^2 \bar{p}^2 n^4}\right).
	\end{equation*}
	Let $j_1 \coloneqq \max\{j: p_j <p^*\}$, $j_2 \coloneqq \min\{j: p_j \ge p^*\}$. Note that $j_2=j_1+1$. By \eqref{equ:mis-unimodal-p} and under event $G$, we have
	\begin{align*}
	&\hat{r}(p_0) \le r(p_0)+ \Delta < r(p_1)-\Delta \le \hat{r}(p_1) <\cdots< \hat{r}(p_{j_1}), \\
	&\hat{r}(p_{j_2}) \ge r(p_{j_2})-\Delta > r(p_{j_2+1})+\Delta \ge \hat{r}(p_{j_2+1}) > \cdots > \hat{r}(p_n).
	\end{align*}
	Thus, the estimated revenue $r(p_j)$ strictly increases first, then strictly decreases. So the optimal index $j^* \in \{j_1,j_2\}$ and $p^* \in [p_{j_1},p_{j_2}] \subset [p_{j^*-1},p_{j^*+1}]$.  \Halmos

\textbf{Proof of Theorem \ref{thm:misspec}.}
	If $\pa \in [p_{j^*-1},p_{j^*+1}]$, then $\prin=\pa$ and $(\prin-c)f(\prin) = (\pa-c)f(\pa)$. If $\pa > p_{j^*+1}$, then $p^* \le \prin=p_{j^*+1}<\pa$ and $(\prin-c)f(\prin) \ge (\pa-c)f(\pa)$, since $(p-c)f(p)$ is unimodal. Also, we have $(\prin-c)f(\prin) \ge (\pa-c)f(\pa)$ when $\pa < p_{j^*-1}$.
	Therefore, we complete the proof. \Halmos

\section{Proofs in Section \ref{sec:conta-LR}}
\textbf{Proof of Lemma \ref{lem:conta-res}.}
	Since there is a constant term in the covariate $W$, we define $W=(1 \  W'^\top)^\top$ and rewrite the true model and the contaminated model in the following:
	\begin{equation*}
	X=
	\begin{pmatrix}
	1 & W'^\top
	\end{pmatrix}
	\begin{pmatrix}
	\beta_0\\
	\beta_1
	\end{pmatrix}
	+ \epsilon, \ X_C=
	\begin{pmatrix}
	1 & W'^\top
	\end{pmatrix}
	\begin{pmatrix}
	\beta_0\\
	\beta_1
	\end{pmatrix}
	+B+\epsilon.
	\end{equation*}
	We rewrite the contaminated model in the matrix form:
	\begin{equation*}
	\bm{X_C}=A \beta+ \bm{B} + \bm{\epsilon},
	\end{equation*}
	where the vector of response, the design matrix, the vector of contamination and noise are
	\begin{equation*}
	\bm{X_C}=
	\begin{pmatrix}
	X_{C1} \\
	\vdots \\
	X_{Cn}
	\end{pmatrix}, \,
	A=
	\begin{pmatrix}
	1 & W_1'^\top\\
	\vdots & \vdots \\
	1 & W_n'^\top
	\end{pmatrix}, \,
	\bm{B}=
	\begin{pmatrix}
	B_1\\
	\vdots \\
	B_n
	\end{pmatrix}, \,
	\bm{\epsilon}=
	\begin{pmatrix}
	\epsilon_1\\
	\vdots \\
	\epsilon_n
	\end{pmatrix}.
	\end{equation*}
	The OLS estimator for the contaminated training samples is 
	\begin{equation}\label{equ:conta-OLS-res}
	\hat{\beta}=(A^\top A)^{-1} A^\top \bm{X_C}=(A^\top A)^{-1} A^\top \left(A \beta + \bm{B}+\bm{\epsilon}\right)=\beta+\left(\frac{1}{n}A^\top A\right)^{-1} \frac{1}{n} A^\top \bm{B} +\left(\frac{1}{n}A^\top A\right)^{-1} \frac{1}{n} A^\top \bm{\epsilon}.
	\end{equation}
	By Corollary 3.1 in \citet{wooldridge2010econometric}, we have 
	\begin{equation}\label{equ:con-inverse-design}
	\left(\frac{1}{n}A^\top A\right)^{-1} \overset{p}{\longrightarrow} \left(\EX[WW^\top]\right)^{-1}=\begin{pmatrix}
	1 & \EX[W']^\top \\
	\EX[W'] & \EX[W' W'^\top]
	\end{pmatrix}^{-1}.
	\end{equation}
	By the inverses of partitioned matrices \citep{greene2003econometric}, we have \eqref{equ:con-inverse-design}
	\begin{equation}\label{equ:con-inverse-result}
	=
	\begin{pmatrix}
	1+\EX[W']^\top \left(\EX[W' W'^\top]-\EX[W'] \EX[W']^\top\right)^{-1} \EX[W'] & -\EX[W']^\top \left(\EX[W' W'^\top]-\EX[W'] \EX[W']^\top\right)^{-1} \\
	-\left(\EX[W' W'^\top]-\EX[W'] \EX[W']^\top\right)^{-1} \EX[W'] & \left(\EX[W' W'^\top]-\EX[W'] \EX[W']^\top\right)^{-1}
	\end{pmatrix}.
	\end{equation}
	For the last term in \eqref{equ:conta-OLS-res}, we have 
	\begin{equation}\label{equ:con-Wepsilon}
	\frac{1}{n} A^\top \epsilon
	=\begin{pmatrix}
	\frac{1}{n} \sum_{t=1}^n \epsilon_t\\
	\frac{1}{n} \sum_{t=1}^n W'_t \epsilon_t
	\end{pmatrix}
	\overset{p}{\longrightarrow} \begin{pmatrix}
	\EX[\epsilon_t]\\
	\EX[W'_t \epsilon_t]
	\end{pmatrix}
	=0,
	\end{equation}
	where $\overset{p}{\longrightarrow}$ denotes convergence in probability, the convergence follows by the weak law of large numbers and the last equality follows by the independence of $\epsilon_t$ and $W'_t$.
	Then, by Slutsky's theorem and \eqref{equ:con-inverse-design}, \eqref{equ:con-inverse-result}, \eqref{equ:con-Wepsilon}, we have
	\begin{equation}\label{equ:con-epsilon-conver}
	\left(\frac{1}{n}A^\top A\right)^{-1} \frac{1}{n} A^\top \bm{\epsilon} 	\overset{p}{\longrightarrow}0.
	\end{equation}
	
	Considering the term including $\bm{B}$ in \eqref{equ:conta-OLS-res}, we have 
	\begin{equation}\label{equ:con-WB}
	\frac{1}{n} A^\top \bm{B}=\begin{pmatrix}
	\frac{1}{n} \sum_{t=1}^n B_t\\
	\frac{1}{n} \sum_{t=1}^n W'_t B_t
	\end{pmatrix}
	\overset{p}{\longrightarrow} \begin{pmatrix}
	\EX[B]\\
	\EX[W'B]
	\end{pmatrix}
	=\begin{pmatrix}
	1\\
	\EX[W']
	\end{pmatrix} \EX[B],
	\end{equation}
	where the last equality follows by the independence of $B_t$ and $W'_t$.
	
	By \eqref{equ:con-inverse-result} and \eqref{equ:con-WB}, we have 
	\begin{equation}\label{equ:con-B-conver}
	\left(\frac{1}{n}A^\top A\right)^{-1} \frac{1}{n} A^\top \bm{B} \overset{p}{\longrightarrow}
	\begin{pmatrix}
	1\\
	\bm{0}
	\end{pmatrix} \EX[B].
	\end{equation}
	Finally, plugging \eqref{equ:con-epsilon-conver}, \eqref{equ:con-B-conver} into \eqref{equ:conta-OLS-res}, we have
	\begin{equation*}
	\hat{\beta} \overset{p}{\longrightarrow}
	\beta+
	\begin{pmatrix}
	1\\
	\bm{0}
	\end{pmatrix} \EX[B], \ \xa(W)=W^\top  \hat{\beta} \overset{p}{\longrightarrow} W^\top \beta+\EX[B]. \Halmos
	\end{equation*} 

\textbf{Proof of Proposition \ref{prop:conta-response}.}
	\begin{figure}[tbp]
		\centering
		\includegraphics[width=0.8\textwidth]{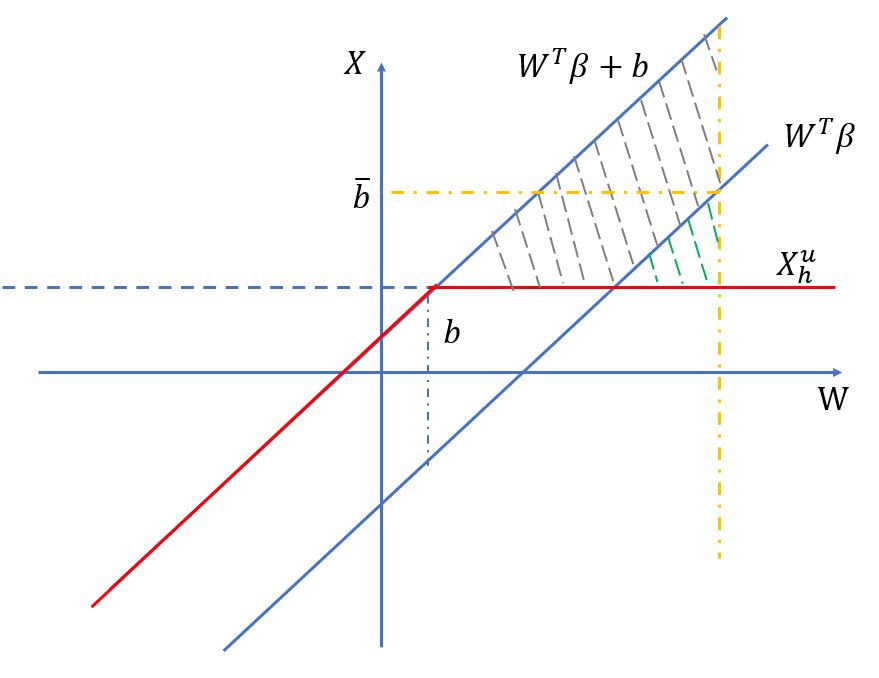}
		\caption{Intuition behind the proof of Proposition \ref{prop:conta-response}.}
	\end{figure}
	Recalling that $\EX[B]=p b$, $\xa(W)=W^\top \beta+p b$ and the loss function is defined as the square error, i.e., $l(\xa(w),w)=(\xa(w)-\xx(w))^2$.
	We write down the difference of the losses,
	\begin{align}\label{equ:SE-dif}
	l(\xa(w),w)-l(\xin(w),w)
	&=(\xa(w)-w^\top \beta )^2 - (\xin(w)-w^\top \beta )^2 \notag\\
	&=\left(p^2 b^2-(\xhu-w^\top \beta)^2 \right)\mathbb{I}(\xhu \leq w^\top \beta +p b).
	\end{align}
	Since $\xhu$ satisfies \eqref{equ:conta-LR-response}, we have 
	\begin{equation}
	w^\top \beta \le \xhu+p b, \ \forall \ w \in \mathcal{W}.
	\end{equation}
	We will show $\eqref{equ:SE-dif} \ge 0$ by discussing $w$ in the following three cases.
	\begin{enumerate}
		\item[(1)] For the $w$ satisfying $\xhu \le w^\top \beta \le \xhu+p b$, we have $0 \le w^\top \beta-\xhu \le p b$. Thus, $p^2 b^2-(\xhu-w^\top \beta)^2 \ge 0$ and $\eqref{equ:SE-dif} \ge 0$.
		\item[(2)] For the $w$ satisfying $\xhu-p b \le w^\top \beta \le \xhu$, we have $-p b \le w^\top \beta -\xhu \le 0$. Thus, $p^2 b^2-(\xhu-w^\top \beta)^2 \ge 0$ and $\eqref{equ:SE-dif} \ge 0$.
		\item[(3)] For the $w$ satisfying
		$w^\top \beta \le \xhu-p b$, we have $\eqref{equ:SE-dif} = 0$.
	\end{enumerate}
	
	We have proved  $\eqref{equ:SE-dif} \ge 0$ for any $w \in \mathcal{W}$. Thus, we have $l(\xa(w),w) \ge l(\xin(w),w)$ for any $w \in \mathcal{W}$. \Halmos

\textbf{Proof of Theorem \ref{thm:conta-respon}.}
	\begin{figure}[tbp]
		\centering
		\includegraphics[width=0.8\textwidth]{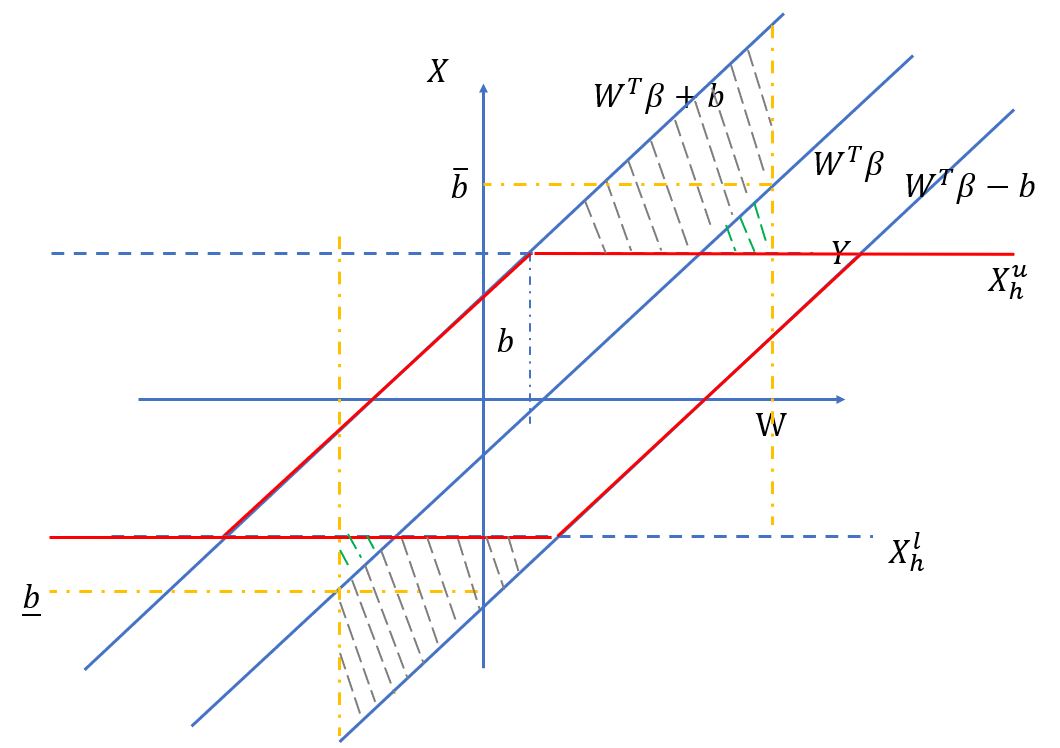}
		\caption{Intuition behind the proof of Theorem \ref{thm:conta-respon}.}
	\end{figure}

	Let $b \coloneqq \EX[B]$, then $\xa(W)=W^\top \beta+b$.
	We first write down the difference of the square-error losses:
	\begin{align}
	&l(\xa(w),w)-l(\xin(w),w) \notag\\
	&=(\xa(w)-w^\top \beta )^2 - (\xin(w)-w^\top \beta )^2 \notag\\
	&\overset{(a)}{=}\left((\xa(w)-w^\top \beta)^2-(\xhu-w^\top \beta)^2\right) \mathbb{I}\left(\xhu \le \xa(w)\right) \notag\\
	&\quad+\left((\xa(w)-w^\top \beta)^2-(\xhl-w^\top \beta)^2\right) \mathbb{I}\left(\xhl \ge \xa(w)\right) \notag\\
	&\overset{(b)}{=}\left(b^2-(\xhu-w^\top \beta)^2\right) \mathbb{I}\left(\xhu \le w^\top \beta+b\right)+\left(b^2-(\xhl-w^\top \beta)^2\right) \mathbb{I}\left(\xhl \ge w^\top \beta+b\right), \label{equ:SE-dif-two}
	\end{align}
	where $(a)$ follows from the definition of $\hat{X}(w)$, $(b)$ follows from $\xa(W)=W^\top \beta+b$.
	Since $\xhl,\xhu$ satisfies \eqref{equ:conta-LR-response-two-side}, we have for any $w \in \mathcal{W}$,
	\begin{align}
	&\xhl-b \le w^\top \beta \le \xhu +b, \ \textrm{if} \ b \ge 0, \label{equ:conta-LR-response-two-side-pr}\\
	&\xhl+b \le w^\top \beta \le \xhu -b, \ \textrm{if} \ b < 0. \label{equ:conta-LR-response-two-side-pr2}
	\end{align}
	We will show $\eqref{equ:SE-dif-two} \ge 0$ by discussing the sign of $b$ and range of $w$ in the following cases.
	\begin{enumerate}
		\item [(1)] If $b\ge 0$, by \eqref{equ:conta-LR-response-two-side-pr}, we have $w^T \beta-b \le \xhu$. For the first term in \eqref{equ:SE-dif-two}, if $\xhu$ further satisfies $\xhu \le w^T \beta+b $, we have $b^2-(\xhu-w^\top \beta)^2 \ge 0$. For the second term in \eqref{equ:SE-dif-two}, by \eqref{equ:conta-LR-response-two-side-pr}, we have $\xhl \le w^T \beta+b$. So we have $\left(b^2-(\xhl-w^\top \beta)^2\right) \mathbb{I}\left(\xhl \ge w^\top \beta+b\right)=0$.
		
		\item[(2)] If $b<0$, we have $\xhu \ge w^\top \beta +b$ due to \eqref{equ:conta-LR-response-two-side-pr2}. Thus, we have 
		\begin{equation*}
		\left(b^2-(\xhu-w^\top \beta)^2\right) \mathbb{I}\left(\xhu \le w^\top \beta+b\right) = 0.
		\end{equation*}
		Next, we will show $\left(b^2-(\xhl-w^\top \beta)^2\right) \mathbb{I}\left(\xhl \ge w^\top \beta+b\right) \ge 0$.
		\begin{enumerate}
			\item [(2.1)] For the $w$ satisfying $\xhl+b \le w^\top \beta \le \xhl$, we have $b \le w^\top \beta-\xhl \le 0$. Thus, $b^2-(\xhl-w^\top \beta)^2 \ge 0$ and $\left(b^2-(\xhl-w^\top \beta)^2\right) \mathbb{I}\left(\xhl \ge w^\top \beta+b\right) \ge 0$.
			\item [(2.2)] For the $w$ satisfying $\xhl \le w^\top \beta \le \xhl-b$, we have $0 \le w^\top \beta-\xhl \le -b$. Thus, $b^2-(\xhl-w^\top \beta)^2 \ge 0$ and $\left(b^2-(\xhl-w^\top \beta)^2\right) \mathbb{I}\left(\xhl \ge w^\top \beta+b\right) \ge 0$.
			\item[(2.3)] For the $w$ satisfying $\xhl-b \le w^\top \beta$, we have $\left(b^2-(\xhl-w^\top \beta)^2\right) \mathbb{I}\left(\xhl \ge w^\top \beta+b\right) = 0$.
		\end{enumerate}
	\end{enumerate}

	Therefore, we have proved  $\eqref{equ:SE-dif-two} \ge 0$ for any $w \in \mathcal{W}$. That is, 
	$l(\xin(w),w) \le l(\xa(w),w)$ for any $w \in \mathcal{W}$.
	Taking expectation of $w$, we have 
	\begin{equation*}
	\EX[l(\xa(W),W)]-\EX[l(\xin(W),W)]=\EX \left[(\xa(W)-W^\top \beta )^2\right] - \EX \left[(\xin(W)-W^\top \beta )^2\right] \ge 0. \Halmos
	\end{equation*}

\textbf{Proof of Lemma \ref{lem:conta-cov}.}
	Recall the observed covariate $W$, contamination error $U$ and the true covariate $Z=W-U$. We have 
	\begin{equation}
	X=Z^\top \beta +\epsilon=W^\top \beta-U^\top \beta+\epsilon.
	\end{equation}
	Suppose the design matrix for $W,U$ are $A,B$. Then the OLS estimator is 
	\begin{equation}\label{equ:conta-cov-OLS}
	\hat{\beta}=(A^\top A)^{-1} A^\top (A \beta-B^\top \beta + \bm{\epsilon})=\beta-(A^\top A)^{-1} A^\top B^\top \beta+(A^\top A)^{-1} A^\top \bm{\epsilon}.
	\end{equation}
	We have 
	\begin{equation}\label{equ:conta-cov-B}
	\frac{1}{n} A^\top A \overset{p}{\longrightarrow} \EX[W W^\top]\overset{(a)}{=}\EX[Z Z^\top]+ \EX[UU^\top]\overset{(b)}{=}\Sigma_1+ \Sigma_2,
	\end{equation}
	where $(a)$ follows from the independence of $W$ and $U$, $(b)$ follows from Assumption \ref{ass:conta-cov-design}.
	Also, we have 
	\begin{equation}\label{equ:conta-cov-eps}
	\frac{1}{n} A^\top B \overset{p}{\longrightarrow}\EX[(U+Z)U^\top]= \EX[UU^\top]= \Sigma_2, \ (A^\top A)^{-1} A^\top \bm{\epsilon} \overset{p}{\longrightarrow} 0.
	\end{equation}
	Plugging \eqref{equ:conta-cov-B} and \eqref{equ:conta-cov-eps} into \eqref{equ:conta-cov-OLS}, we have 
	\begin{equation*}
	\hat{\beta} \overset{p}{\longrightarrow} \left(\mathcal{I}- (\Sigma_1+ \Sigma_2)^{-1} \Sigma_2\right) \beta.
	\end{equation*}
	Since $\Sigma_1$ is positive-definite and $\Sigma_2$ is positive semi-definite, $ (\Sigma_1+p \Sigma_2)^{-1} \Sigma_2\beta=0$ if and only if $\Sigma_2 \beta=0$. \Halmos

\textbf{Proof of Theorem \ref{thm:conta-covariates}.}
	Recall that the OLS estimator $\hat{\beta}=\beta$ and $\xa(Z)=Z^\top \beta+U^\top \beta$.
	Note that $Z$ is independent of $U$. We first fix a covariate $z$ and reformulate $\EX[l(\xa(z),z)]-\EX[l(\xin(z),z)]$, where the expectation is taken with respect to $U$. That is 
	\begin{align}
	&\EX[l(\xa(z),z)]-\EX[l(\xin(z),z)] \notag\\
	&=\EX\left[(\xa(z)-z^\top \beta )^2\right] - \EX\left[(\xin(z)-z^\top \beta )^2\right] \notag\\
	&\overset{(a)}{=}\EX\left[\left((\xa(z)-z^\top \beta)^2-(\xhu-z^\top \beta)^2\right) \mathbb{I}\left(\xhu \le \xa(z)\right)\right] \notag\\
	&\quad+\EX\left[\left((\xa(z)-z^\top \beta)^2-(\xhl-z^\top \beta)^2\right) \mathbb{I}\left(\xhl \ge \xa(z)\right)\right] \notag\\
	&\overset{(b)}{=}\EX\left[\left((U^\top \beta)^2-(\xhu-z^\top \beta)^2\right) \mathbb{I}(\xhu \le z^\top \beta+U^\top \beta)\right] \notag\\
	&\ \ +\EX\left[\left((U^\top \beta)^2-(\xhl-z^\top \beta)^2\right) \mathbb{I}(\xhl \ge z^\top \beta+U^\top \beta)\right]\notag\\
	&=\EX\left[\left(B^2-\rup^2(z)\right) \mathbb{I}(\rup(z) \le B) \right]+\EX\left[\left(B^2-\rlw^2(z)\right) \mathbb{I}(\rlw(z) \ge B) \right],
	\label{equ:SE-dif-two-cov}
	\end{align}
	where 
	\begin{equation}\label{equ:LR-conta-cov-u}
	B \coloneqq U^\top \beta, \ \rup(z) \coloneqq \xhu-z^\top \beta, \ \rlw(z) \coloneqq \xhl-z^\top \beta.
	\end{equation}
	Furthermore, $(a),(b)$ follow from the definition of $\xin,\xa$.
	According to \eqref{equ:LR-conta-cov-v}, we have 
	\begin{equation}\label{equ:LR-conta-cov-B}
	\PR(B \ge b) \ge p, \ \PR(B \le -b) \ge p.
	\end{equation}
	According to \eqref{equ:conta-cov-discrete} and $p\le 0.5$, we have
	\begin{equation}\label{equ:LR-conta-cov-range}
	b \overset{(a)}{\ge} \sqrt{\frac{p}{1-p}}b \overset{(b)}{\ge} \max_{z \in \mathcal{Z}} \{z^\top \beta\}- \xhu \overset{(c)}{\ge} -\rup(z), \ -b \le -\sqrt{\frac{p}{1-p}}b \le \min_{z \in \mathcal{Z}} \{z^\top \beta\}- \xhl \le -\rlw(z),
	\ \forall z \in \mathcal{Z},
	\end{equation}
	where $(a)$ follows from $p \le 0.5$, $(b)$ follows from \eqref{equ:conta-cov-discrete}, $(c)$ follows from \eqref{equ:LR-conta-cov-u}. 
	Next, we will show $\eqref{equ:SE-dif-two-cov} \ge 0$ by discussing the range of $z$. The second series of inequality can be derived similarly.
	\begin{enumerate}
		\item [(1)] For the $z$ satisfying $z^\top \beta \ge \xhu$, we have $\rup(z) \le 0$ and $\rlw(z) \le 0$ by \eqref{equ:LR-conta-cov-u}. Thus, we have
		\begin{equation}\label{equ:LR-conta-cov-lob}
		\EX\left[\left(B^2-\rlw^2(z)\right) \mathbb{I}(\rlw(z) \ge B) \right] \ge 0,
		\end{equation}  
		and 
		\begin{align}
		&\EX\left[\left(B^2-\rup^2(z)\right) \mathbb{I}(\rup(z) \le B) \right]\notag\\
		& \overset{(a)}{=} \PR(B \ge b) \left((B^2-\rup(z)^2) \mathbb{I}(\rup(z) \le B)\right)+\PR(-b \le B \le b) \left(B^2-\rup^2(z)\right) \mathbb{I}(\rup(z) \le B) \notag\\
		& \overset{(b)}{\ge} \PR(B \ge b) \left((b^2-\rup(z)^2) \mathbb{I}(\rup(z) \le b)\right)+\PR(-b \le B \le b) \left(B^2-\rup^2(z)\right) \mathbb{I}(\rup(z) \le B) \notag\\
		& \overset{(c)}{\ge} \PR(B \ge b) \left((b^2-\rup(z)^2) \mathbb{I}(\rup(z) \le b)\right)-\left(\PR(-b \le B \le b)\right) \rup^2(z) \mathbb{I}(\rup(z) \le 0) \notag\\
		& \overset{(d)}{\ge} p (b^2-\rup^2(z))-(1-2p) \rup^2(z) \notag\\
		&=p b^2-(1-p) \rup^2(z) \notag\\
		&=p_1 b^2-(1-p) (\xhu-z^\top \beta)^2 \notag\\
		& \overset{(e)}{\ge} 0, \label{equ:LR-conta-cov-upb}
		\end{align}
		where $(a)$ follows by $\PR(B < -b)=0$ due to $b \ge -\rup(z) \ge -B$ by \eqref{equ:LR-conta-cov-range},
		$(b)$ holds by $(B^2-\rup(z)^2) \mathbb{I}(\rup(z) \le B)$ increasing in $B$ when $B \ge b$,
		$(c)$ holds by $\left(B^2-\rup^2(z)\right) \mathbb{I}(\rup(z) \le B) \ge -\rup^2(z) \mathbb{I}(\rup(z) \le 0)$ when $-b \le B \le b$ due to $-b \le \rup(z) \le 0$ by \eqref{equ:LR-conta-cov-range}, $(d)$ follows by \eqref{equ:LR-conta-cov-B} and  $\rup(z) \le 0 \le -\rup(z)\le b$ by \eqref{equ:LR-conta-cov-range}, $(e)$ follows by \eqref{equ:conta-cov-discrete}.
		Plugging \eqref{equ:LR-conta-cov-lob} and \eqref{equ:LR-conta-cov-upb} into \eqref{equ:SE-dif-two-cov}, we have $\eqref{equ:SE-dif-two-cov} \ge 0$.
		\item[(2)] For the $z$ satisfying $\xhl \le z^\top \beta \le \xhu$, we have $\rup(z) \ge 0$ and $\rlw(z) \le 0$ by  \eqref{equ:LR-conta-cov-u}. Thus, we have $\EX\left[\left(B^2-\rup^2(z)\right) \mathbb{I}(\rup(z) \le B) \right] \ge 0$ and $\EX\left[\left(B^2-\rlw^2(z)\right) \mathbb{I}(\rlw(z) \ge B) \right] \ge 0$. Then, $\eqref{equ:SE-dif-two-cov} \ge 0$.
		\item[(3)] For the $z$ satisfying $z^\top \beta \le \xhl$, we have $\rup(z) \ge 0$ and $\rlw(z) \ge 0$ by  \eqref{equ:LR-conta-cov-u}. Thus, we have 
		\begin{equation}\label{equ:LR-conta-cov-upb-l}
		\EX\left[\left(B^2-\rup^2(z)\right) \mathbb{I}(\rup(z) \le B) \right] \ge 0,
		\end{equation}
		and 
		\begin{align}
		&\EX\left[\left(B^2-\rlw^2(z)\right) \mathbb{I}(\rlw(z) \ge B) \right] \notag\\
		& \overset{(a)}{=} \PR(B \le -b) \left((B^2-\rlw(z)^2) \mathbb{I}(\rlw(z) \ge B)\right)
		-\PR(-b \le B \le b)	\left(B^2-\rlw^2(z)\right) \mathbb{I}(\rlw(z) \ge B)	\notag\\
		& \overset{(b)}{\ge} \PR(B \le -b) \left((b^2-\rlw(z)^2) \mathbb{I}(\rlw(z) \ge -b)\right)-\left(\PR(-b \le B \le b)\right) \rlw^2(z) \mathbb{I}(\rup(z) \ge 0) \notag\\
		& \overset{(c)}{\ge} p (b^2-\rlw^2(z))-(1-2p) \rlw^2(z) \notag\\
		& = p b^2-(1-p)\rlw^2(z)\notag\\
		& = p b^2-(1-p) (\xhl-z^\top \beta)^2 \notag\\
		& \overset{(d)}{\ge} 0, \label{equ:LR-conta-cov-upb-u}
		\end{align}
		where $(a)$ follows by $\PR(B > b) = 0$ due to $-b \le -\rlw(z) \le -B$ by \eqref{equ:LR-conta-cov-range},
		$(b)$ follows by $(B^2-\rlw(z)^2) \mathbb{I}(\rlw(z) \ge -B)$ decreasing in $B$ when $B \le -b$ and  $\left(B^2-\rlw^2(z)\right) \mathbb{I}(\rlw(z) \ge B) \ge -\rlw^2(z) \mathbb{I}(\rlw(z) \ge 0)$ when $-b \le B \le b$ due to $0 \le \rlw(z) \le b$ by \eqref{equ:LR-conta-cov-range}, $(c)$ follows by \eqref{equ:LR-conta-cov-B} and $\rlw(z) \ge 0 \ge -b$, $(d)$ follows by \eqref{equ:conta-cov-discrete}. Plugging \eqref{equ:LR-conta-cov-upb-l},  \eqref{equ:LR-conta-cov-upb-u} into \eqref{equ:SE-dif-two-cov}, we have $\eqref{equ:SE-dif-two-cov} \ge 0$.
	\end{enumerate}
	Therefore, we have proved $\eqref{equ:SE-dif-two-cov} \ge 0$ for any $z \in \mathcal{Z}$. Finally, taking expectation with respect to $Z$, we have 
	\begin{equation*}
	\EX[l(\xa(W),Z)]-\EX[l(\xin(W),Z)]=\EX\left[(\xa(W)-Z^\top \beta )^2\right] - \EX\left[(\xin(W)-Z^\top \beta )^2\right] \ge 0. \Halmos
	\end{equation*}

\end{document}